\documentclass{report}

\usepackage{arxiv}

\usepackage{natbib}
\usepackage{amsmath}
\usepackage[export]{adjustbox}
\usepackage{dsfont}
\usepackage{tikz-cd}
\usepackage{amsthm}
\usepackage{subfig}
\usepackage{amsfonts}
\usepackage{amssymb}

\DeclareMathAlphabet{\mathcal}{OMS}{cmsy}{m}{n}

\DeclareMathOperator{\ext}{ext}
\DeclareMathOperator{\vc}{VC}

\DeclareMathOperator{\comp}{comp}
\DeclareMathOperator{\kl}{KL}
\DeclareMathOperator{\hinge}{hinge}

\DeclareMathOperator{\trace}{trace}
\DeclareMathOperator{\tRad}{tRad}
\DeclareMathOperator{\empRad}{empRad}
\DeclareMathOperator{\sign}{sign}

\newtheorem{conjecture}{Conjecture}
\newtheorem{theorem}{Theorem}
\newtheorem{corollary}{Corollary}
\newtheorem{definition}{Definition}

\title{Improvability through Semi-Supervised Learning:\\ A Survey of Theoretical Results}

\author{
   Alexander Mey \\
  Delft University of Technology, The Netherlands\\
  \texttt{a.mey@tudelft.nl} \\
   \And
   Marco Loog\\
  Delft University of Technology, The Netherlands\\
University of Copenhagen, Denmark \\
  \texttt{m.loog@tudelft.nl} \\
}

\begin{document}

\title{Semi-Supervised Learning\\ and Improved Generalization:\\ A Survey of Theoretical Results}

\author{
   Alexander Mey \\
  Delft University of Technology, The Netherlands\\
  \texttt{a.mey@tudelft.nl} \\
   \And
   Marco Loog\\
  Delft University of Technology, The Netherlands\\
University of Copenhagen, Denmark \\
  \texttt{m.loog@tudelft.nl} \\
}

\maketitle

\begin{abstract}
Semi-supervised learning is the learning setting in which we have both labeled and unlabeled data at our disposal. In this survey, we explore different types of theoretical results for this setting and map out the benefits of unlabeled data in classification and regression tasks. Most methods that use unlabeled data rely on certain assumptions about the data distribution. When those assumptions are not met in reality, including unlabeled data may actually decrease performance. When applying such methods, it is particularly instructive therefore to have an understanding of the underlying theory and the possible learning behavior that comes with it. In this review, we gather results about the possible gains one can achieve when using semi-supervised learning as well as results about the limits of such methods. More precisely, this review collects the answers to the following questions: what are, in terms of improving supervised methods, the limits of semi-supervised learning? What are the assumptions of different methods? What can we achieve if the assumptions are true? Finally, we also discuss the biggest bottleneck of semi-supervised learning, namely the assumptions they make.
\end{abstract}

\chapter{Introduction}

For various applications, gathering unlabeled data is easier, faster, and/or cheaper than gathering labeled data. The goal of semi-supervised learning (SSL) is to combine unlabeled and labeled data so as to design classification or regression rules that outperform schemes that are only based on the labeled data. SSL does come, however, with an inherent risk. It is well known that including unlabeled data can degrade the performance \citep{BenDavid,risks}.

Studying and understanding SSL from a theoretical point of view allows us to exactly formulate the assumptions we need, the improvements we can expect, as well as the limitations of the different methods. Based on such understanding, one can then formulate recommendations for using SSL with the aim of avoiding any decrease in performance as good as possible. In this review, we collect and discuss such theoretical results concerning SSL, study the relevant papers in detail, present their main result, and point out connections between these papers and connections to other work.

Our review paper tries to target two groups of audience. The first group consists of interested practitioners and researchers working on experimental SSL and its applications. While they may not be interested in all the details we present, the introduction to each of the chapters should give a good high-level understanding of the types of theoretical results in this field and the main insights they provide. The other target audience is anybody working on the theoretical side of SSL. We are convinced that, especially researchers starting in this field, will find inspiration and connections to their own work in our overview.

We mostly present results that describe the performance of semi-supervised learners, often, but not exclusively, in the language of the PAC-learning framework.\footnote{PAC-learning stands for \emph{Probabilistically Approximately Correct}-learning. In this framework one can study how far a trained classifier is off of the best classifier from a given class, given a certain amount of labeled data. The rate at which we approach the best classifier is called learning rate. Good introductions to this framework can be found in \citep{Shalev} and \citep{Mohri}. We also refer to Definition \ref{supsampcomp}, where we introduce the notion of sample complexity. PAC-learnable means that the sample complexity is always finite.} We interpret the results, draw connections between them and point out what one has to assume for them to be valid. Next to theoretical guarantees of some specific learners, we also present results on the limits of SSL.

\section{Outline}
In Chapter \ref{Impossibility} we discuss results on the limits of SSL, which typically arise due to specific assumptions about the model or the data generation process. We also discuss the limits of semi-supervised methods that do not make any assumption about the data distribution. As opposed to the settings where the improvements of SSL are provably limited, we present in this same chapter three settings where the improvements of SSL are \emph{unlimited}. With unlimited we mean here that a semi-supervised learner can PAC-learn the problem, while no supervised learner (SL) can.

In Chapter \ref{Noassumptions}, we investigate what is possible with some specific methods that try to exploit unlabeled data without making further assumptions on the data distribution.
In Chapter \ref{weakassumptions}, we treat semi-supervised learners that make \emph{weak} assumptions on the data distribution. The assumptions are weak in the sense that the resulting learner cannot get a learning rate faster than the standard learning rate of $\frac{1}{\sqrt{n}}$,\footnote{The learning rate is the rate in which we converge to the best classifier from a given class in number of the labeled samples. That the standard rate (without further assumptions) is in order of $\frac{1}{\sqrt{n}}$ follows from classic results as shown for example by \citet{Vapnik1998}. Similar results can be found in the works mentioned in the previous footnote.} where $n$ is the number of labeled samples. Instead, the improvements are given by a constant

In Chapter \ref{strongassumptions}, we discuss learners that use \emph{strong} assumptions under which one can converge exponentially fast to the best classifier in a given class, i.e., the learning rate is in order of $e^{-n}$. We remark that there is not necessarily a principled qualitative difference in weak and strong assumptions, but rather a subtle quantitative difference. Both might make assumptions about the data distribution, but strong assumptions are just that, stronger. We will elaborate on that in the beginning of Chapter \ref{strongassumptions}.

In Chapter \ref{Transductive}, we present results in the transductive setting, a setting where one is only interested in the labels of the unlabeled data available. In the same chapter, we also present a line of research that tries to construct semi-supervised learners that are never worse than their supervised counterparts.

In the final chapter, Chapter \ref{discussion}, we discuss the overall results and conclude with what we see as the current challenges in the field.

Before turning to the next chapter, however, we first introduce the formal learning framework that is assumed in the majority of the works surveyed.

\section{The Learning Framework} \label{framework}

Unless further specified, all results are presented in the standard statistical learning framework. This means that we are given a feature space $\mathcal{X}$ and a output space $\mathcal{Y}$ together with an unknown distribution $P$ on $\mathcal{X} \times \mathcal{Y}$. For the most part of the survey we will work in the classification setting, such that $\mathcal{Y}=\{-1,1\}$. With slight abuse of notation, we write $P(X)$ and $P(Y)$ for the marginal distributions on $\mathcal{X}$ and $\mathcal{Y}$. Similar conventions are used for conditional distributions.

We consider the setting in which we have observed a labeled $n$-sample
\[
S_n=((x_1,y_1),\ldots,(x_n,y_n))
\]
and an unlabeled $m$-sample
\[
U_m=(x_{n+1},\ldots,x_{n+m}),
\]
where each $(x_i,y_i)$ for $1\leq i \leq n$ and each $x_j$ for $n+1 \leq j \leq n+m$ is identically and independently distributed according to $P$. One then chooses a hypothesis class $H$, where each $h \in H$ is a mapping
\[
h: \mathcal{X} \to \mathcal{Y},
\]
and a loss function
\[
l: \mathcal{Y} \times \mathcal{Y} \to \mathbb{R}.
\]
Unless specified otherwise, we assume for classification that $\mathcal{Y}=\{-1,+1\}$ and the loss is the 0-1 loss:
\[
l(y,\hat{y})=I_{\{y \neq \hat{y}\}}.
\]
For the regression task, we assume that $\mathcal{Y}=\mathbb{R}$ and consider the standard squared loss:
\[
l(y,\hat{y})=(y-\hat{y})^2.
\]
Based on the $n$ labeled and $m$ unlabeled samples we then try to find a $h \in H$ such that the risk
\[
R(h):= \mathbb{E}_{X,Y} \left[ l(h(X),Y) \right]
\]
is small.

Whenever we have any quantity $A$ that depends on the distribution $P$, we write $\hat{A}$ for a empirically estimated version of $A$. For example, given a labeled sample $S_n$, we write
\[
\hat{R}(h)=\frac{1}{n} \sum_{i=1}^n l(h(x_i),y_i)
\]
for the empirical risk of $h \in H$ measured on $S_n$. In case it is not clear from the context, the context clarifies on which sample we measure the loss.

In Table \ref{fulllist} on page \pageref{fulllist}, we present a complete list of the notation we employ throughout this survey.

\section{Assumptions in Semi-Supervised Learning} \label{ssassumptions}
Many semi-supervised learning methods rely on assumptions about the data distribution. Following that many results we present will formalize those assumptions in order to state their results. In this section we introduce the most common assumptions and clarify their relation.  

One of most used assumptions is the \emph{semi-supervised smoothness assumption} \cite[Section 1.2]{Chapelle}. This assumption roughly states that two input points that are close together, have a high likelihood to share the same output. The importance here lies in the word \emph{close}. Naively, one could call two points close, when their euclidean distance is small, but in particular when we have unlabeled data one can think about more sophisticated ways for a definition of closeness. 

One angle to approach this is captured in the \emph{cluster assumption}. The idea is that we could use the unlabeled data to find clusters in our data, and then call two points close if they are in the same cluster. Work that formalizes this idea can be found in Section \ref{strongassumptions} and we will see that this assumption is very strong as we can learn exponentially fast with it.

The \emph{low-density separation} can be seen as a specific instance of the cluster assumption, but can give rise to different algorithms. It states that the decision boundary should lie in a region with low-density. Indeed, if we define clusters as regions of high density and would like to separate those, the decision boundary should automatically be in a low-density region. Again, the unlabeled data helps here as with this we can actually identify the low-density regions. We are not aware of any work that formalizes this concept directly and looks at any type of guarantees.

The \emph{manifold assumption} can be related to the above concepts, but has lead to confusion in the past as there are two alternative definitions. The first definition is best explained with a quote from \cite{Belkin}. 'We will assume that if two points $x_1,x_2 \in X$ are close in the intrinsic geometry of $P(X)$, then the conditional distributions $P(y \mid x_1)$ and $P(y \mid x_2)$ are similar'. The manifold refers to the 'intrinsic geometry of $P(X)$' and we again see that is essentially the same concept as the cluster assumption, once we define a cluster as a collection of points from $X$ that are close on the manifold.

The confusion about the manifold assumption stems now from an alternative definition, as for example given in \cite[Section 1.2.3]{Chapelle}. 'The (high-dimensional) data lie (roughly) on a low dimensional manifold.' Note that this definition of the manifold assumption does not assume anything about the distribution $P(Y \mid X)$. Although a low dimensional manifold can help to avoid the curse of dimensionality, an analysis from Section \ref{onlymanifold} reveals that the knowledge of the manifold does not bring any additional advantage regarding minimax rates. 

If not stated otherwise, we will use the first definition of manifold assumption.

To our knowledge that are few assumptions that really diverge from the above concepts, with the notable exception of the multi-view assumption. This assumption essentially states that one can split the feature space into two subspaces, and each subspace is already sufficient to solve the learning problem. We cover one formalization of this in Section \ref{multiview}, where we also explain the intuition how this assumption can actually help in guiding the learning process.

\chapter[Possibility and Impossibility of Semi-Supervised Learning]{Possibility and Impossibility\\ of Semi-Supervised Learning}\label{Impossibility}

In SSL, we want to use information about the distribution on $\mathcal{X}$ to improve learning. It is not necessarily clear, however, that this information can be useful at all. Various works formalize the idea of using unlabeled data and subsequently investigate situations where unlabeled data cannot help or where it, in fact, can. This chapter follows the same division between possibility and impossibility.  In Section \ref{impossibilitysub}, we present different settings for which it has been shown that unlabeled data cannot help, while in Section \ref{effectivessl}, we present three specific settings where unlabeled data can give unlimited improvements. In this context, unlimited means that no supervised learner can PAC-learn in the situation considered, whereas some semi-supervised learner can.

The negative results often assert an independence between the posterior probability $P(Y|X)$ and the marginal distribution $P(X)$. This does, however, not directly mean that unlabeled data is useless, as we are usually not only interested in $P(Y|X)$ but in the complete risk $\mathbb{E}_{X,Y} \left[ l(h(X),Y) \right]$ of a classifier $h$, which \emph{does} depend on $P(X)$ \citep[Subsection 5.1.2]{Peters}. In Sections \ref{sokolovska} and \ref{kari}, for example, we present work that shows risk improvements even when $P(Y|X)$ and $P(X)$ are independent.

\section{Impossibility Results}\label{impossibilitysub}
In this section we present results that describe in different ways the limits of semi-supervised learning methods. Many results show that semi-supervised learning is in a way inherently impossible in some settings. Next to this we present a line of research that investigates the limits of semi-supervised learning methods, when no particular assumptions about the data distribution (as explained in Section \ref{ssassumptions}) are made.

\subsection{Impossibility Because of the Data Generation Process}\label{seeger}
\begin{figure}[ht]
 \centering
 \begin{tikzcd}
\mu \arrow{d} & \theta \arrow{d} \\%
X \arrow{r} & Y
\end{tikzcd}
\caption{The data generation process used in the analysis of Seeger}
\label{Datagen}
\end{figure}
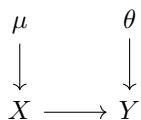
\citet{Seeger} looks at a simple data generation model and investigates how prior information about the data distribution changes our posterior belief about the model if the prior information is included in a Bayesian fashion. To use the Bayesian approach, the data is assumed to be generated in the following manner. To start with, it is assumed that the distribution $P$ comes from a model class with parameters $\mu$ and $\theta$. Subsequently, values $\mu \sim P_\mu$ and $\theta \sim P_\theta$ are sampled independently after which the data is generated by gathering samples $x \sim P(X | \mu)$ with corresponding labels $y \sim P(Y | X,\theta)$. The graphical model for this generation process is given in Figure \ref{Datagen}.

The goal in this setting is to infer $\theta$ from a finite labeled sample $S_n=(x_i,y_i)_{1\leq i \leq n}.$
Using a Bayesian approach it can be easily shown that $P(\theta | S_n)$ is independent of any finite unlabeled sample and $\mu$ itself. In other words: unlabeled information does not change the posterior belief about $\theta$ given the labeled data $S_n$. A possible solution is to assume a dependency between $\mu$ and $\theta$, which comes down to an additional arrow between $\mu$ and $\theta$ in Figure \ref{Datagen}. This exact approach was chosen in Example 1 by \cite{gopfert} to create a setting where knowledge of the marginal distribution can indeed help in the following sense. In their example the marginal distribution completely determines the Bayes classifier, and there exists thus a semi-supervised learner that always has zero risk, while any supervised learner has the standard learning rate of $\frac{1}{\sqrt{n}}$. Alternatively we can of course also think about settings, where the data generation process from Figure \ref{Datagen} is flipped: We first sample a label $y$, and then sample a feature $x$ from a marginal distribution associated to $y$. This setup was chosen in the work we present in Section \ref{exponential}, where unlabeled data was shown to be useful.

\subsection{Impossibility Because of Model Assumptions} \label{hansen}

\citet{Hansen} investigates when unlabeled data should change our posterior belief about a model. In comparison to \citet{Seeger}, no data generation assumptions are made, but rather assumptions about the model that is used. The author looks at solutions derived from the expected squared loss between this given model and the true desired label output. Splitting the joint distribution $P(X,Y | \theta)$ of the model considered as
\[
P(X,Y | \theta)=P(Y|X,\theta_1,\theta_2) P(X|\theta_2,\theta_3),
\]
the conclusion is reached that unlabeled data can be discarded if $\theta_2$, the shared parameter between the label and marginal distribution, is empty. 

In reverse, the effectiveness of methods like expectation maximization \cite{EM} or the provable improvements of the method from Section \ref{mcpl} stem from the fact that some generative models cannot be decomposed in the above way. Assume for example that our data is distributed as two gaussian distributions, where each distribution corresponds to a label. That means that $\theta=\{q,\mu_1,\mu_2,\Sigma_1,\Sigma_2\}$, where $\mu_i$ and $\Sigma_i$, $i \in \{0,1\}$, are the class means and covariance matrices, and $q \in [0,1]$ is the class prior. In that case $P(Y|X,\theta)$ and $P(X|\theta)$ both depend on the class means and covariances.

Earlier work by \citet{Zhang} distinguishes the same type of models, but the impossibility is about the asymptotic efficiency of semi-supervised classifiers. Specifically, the paper considers the following two joint probabilities, which both provide generative models:
\begin{enumerate}
\item Parametric: $P(X,Y| \alpha)=P(X| \alpha)P(Y|X,\alpha)$,
\item Semi-Parametric: $P(X,Y| \alpha)=P(X)P(Y|X,\alpha)$.
\end{enumerate}
In addition, the author shows that the Fisher information $I(\hat{\alpha})_\text{unlab + lab}$ of an maximum likelihood estimator (MLE) $\hat{\alpha}$ that takes labeled an unlabeled data into account can be decomposed as
\[
I(\hat{\alpha})_\text{unlab +lab}=I(\hat{\alpha})_\text{unlab}+I(\hat{\alpha})_\text{lab}.
\]
So, as long as unlabeled data is available, the Fisher information of the semi-supervised learner is bigger compared to the supervised learner, as the latter equals $I(\hat{\alpha})_\text{lab}$. It follows that the semi-supervised learner is asymptotically more efficient, although not necessarily strictly. In the parametric case, we observe that $I(\hat{\alpha})_\text{unlab}=0$ and the semi-supervised and supervised estimator have the same asymptotic behavior. The latter is similar in setting and result to the previous subsection,the biggest difference being that previously we had an impossibility of Bayes updating, while here we have an impossibility of gain in fisher information.

The models considered in this subsection are generative. Section \ref{sokolovska} considers a method that allows for the asymptotic efficiency of a semi-supervised learner even when using a discriminative model $P(Y|X, \alpha)$.

\subsection{Impossibility Because of Causal Direction}

\begin{figure}[ht]
 \centering
 \begin{tikzcd}
N_c \arrow{d} & N_e \arrow{d} \\%
C \arrow{r}{\varrho} & E
\end{tikzcd}
\caption{Simple functional causal model used by \citet{Schoelkopf}. The effect $E$ is caused by $C$ given a deterministic mapping $\varrho$. Both $E$ and $C$ are influenced by a noise variables $N_E$ and $N_C$}.
\label{acausal}
\end{figure}
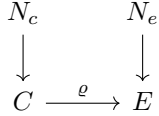

\citet[Sections 2 and 3]{Schoelkopf} analyze a functional causal model such as the one shown in Figure \ref{acausal}. They consider different learning scenarios under the assumption that the label is the cause $C$ and the feature is the effect $E$ and vise versa. This model introduces an asymmetry in cause and effect, since it leads to the fact that $P(C)$ and $P(E|C)$ are independent, while $P(E)$ and $P(C|E)$ are not.  Assuming now that $X$ is the cause of the label $Y$, we find that the prediction $P(Y|X)$ is independent of newly gained information about $P(X)$. This independence vanishes though if we assume that the label $Y$ was caused by $X$. 

One problem, of course, is that we do not necessarily know if the feature is a cause or an effect. But for example in medical settings, this might not be too difficult, as we can identify causal features as those that do actually cause an illness, while effect features are the symptoms of an illness. The work of \citet{Julius} uses such knowledge and derive a SSL method, which only takes the unlabeled data of effect features into account.

\subsection{Impossibility to Always Outperform a Supervised Learner} \label{Jessepaper}

Inspired by a successful approach for a \emph{generative} linear discriminant model by \citet{Loog} (treated in Section \ref{mcpl}), \citet{Jesse} investigate a similar approach to find semi-supervised solutions for \emph{discriminative} models that are never worse than their supervised counterparts. They use a setting where the discriminative models are derived with a monotonously decreasing loss function. The setting is also transductive, so where one is only interested in the performance of the model on the unlabeled data $U_m$, a setting which we discuss in more detail in Section \ref{Transductive}. This work essentially shows that, under some mild conditions, there is always a labeling of the unseen data $U_m$ such that a semi-supervised learner performs worse on $U_m$ than the supervised solution does. In this sense, it is impossible to guarantee that the semi-supervised solution always outperforms the supervised solution.

\subsection{Impossibility If We Only Know the Manifold} \label{onlymanifold}

\citet[Section 3]{Lafferty} shows that knowledge of the manifold alone, without additional assumption, is not sufficient to outperform a purely supervised learner, compare also to Section \ref{ssassumptions} where we explain the manifold assumption of semi-supervised learning. They work in a regression setting and extend work by \citet{Bickel}, who introduced a supervised learning that performs regression on an unknown manifold, to show that there is a supervised learner that can adapt to the dimension of the manifold and thus can achieve minimax rates equivalent to a learner that directly works on the lower dimensional manifold. 

Note that \citet{Lafferty} also show that one can achieve essentially faster rates if we also assume a semi-supervised smoothness assumption. We do not cover more of the relevant details at this point, but offer a qualitatively very similar analysis in Section \ref{densitysensitive}.

\subsection{Impossibility If We Don't Make Additional Assumptions} \label{constantimprove}

\citet{BenDavid} provided a series of investigations starting from the conjecture that SSL is, in some sense, generally not possible without any assumptions. They hypothesize that, at least without any additional distributional assumptions as the ones explained in Section \ref{ssassumptions}, a semi-supervised learner cannot have essentially better sample complexity bounds than a SL (see Definitions \ref{supsampcomp} and \ref{semsampcomp}). This is essentially different from the previous sections, as now there are no further restrictions on the model or the data generation process.

In the following two sections, we want to illustrate the precise idea of these conjectures.  Additionally, we clarify why they do not hold generally and in which scenarios they are generally true.  We start, however, with the main contributions from \citet{BenDavid}.

The generic hypothesis is that the worst-case sample complexity for any semi-supervised learner improves over a supervised learner at most by a constant that only depends on the hypothesis class. The first conjecture states this for the realizable case.
\begin{conjecture}[{\citet[Conjecture 4]{BenDavid}}] \label{bendavid1}
For any hypothesis class $H$, there exists a constant $c(H)$ such that for any domain distribution $D$ on $\mathcal{X}$
\begin{equation}
\sup\limits_{h \in H} m(H,D_h,\epsilon,\delta) \leq \sup\limits_{h \in H} c(H) m^{\text{\normalfont SSL}}(H,D_h,\epsilon,\delta),
\end{equation}
for $\epsilon$ and $\delta$ small enough, where $D_h$ is the distribution on $\mathcal{X} \times \mathcal{Y}$ with marginal distribution $D$ and conditional distribution
\[
D_h(Y=h(x)|X=x)=1.
\]
\end{conjecture}
The second conjecture states the same for the agnostic case, i.e., where we replace $D_h$ for any arbitrary distribution $P$.
\begin{conjecture}[{\citet[Conjecture 5]{BenDavid}}]\label{bendavid2}
For any hypothesis class $H$, there exists a constant $c(H)$ such that for any domain distribution $D$
\begin{equation}
\sup\limits_{P \in \ext(D)} m(H,P,\epsilon,\delta) \leq \sup\limits_{P \in \ext(D)} c(H) m^{\text{\normalfont SSL}}(H,P,\epsilon,\delta),
\end{equation}
for $\epsilon$ and $\delta$ small enough and where $\ext(D)$ is the set of all distributions $P$ on $\mathcal{X} \times \mathcal{Y}$ such that the marginal distribution fulfills $P(X)=D$.
\end{conjecture}
In other words, the paper conjectures that if we are given a fixed domain distribution, one can always find a labeling function for it such that the sample complexity gap between SL and SSL can only be a constant. The paper proofs these conjectures for smooth distributions on the real line and threshold functions in the realizable case and for threshold functions and unions of intervals in the agnostic case.

Here, we already note that the sample complexity comparison is, by construction, a worst case analysis. This means that in cases where the target hypothesis behaves benign, we may still get non-constant improvements. Those cases are further explored in Chapter \ref{strongassumptions}. On another note, we can also ask the question how good a constant improvement by itself can already be. We elaborate on this in the discussion chapter.

The Conjectures \ref{bendavid1} and \ref{bendavid2} are essentially true in the realizable case when the hypothesis class has finite VC-dimension.
\citet{Darnstadt} show that Conjecture \ref{bendavid1}, the realizable case, is true with a small alteration: the supervised learner is allowed to be twice as inaccurate, which is captured by the $2\epsilon$ in Inequality \eqref{darneq1}, and for the finite VC-dimension case we get an additional term of $\log(\frac{1}{\epsilon})$ in Inequality \eqref{darneq2}. \citet{Mey} take this idea a step further and show that there is a setting in which manifold regularization, which uses the manifold assumption, obeys the limits stated by the conjecture, even though in this case the domain distribution carries information about the labeling function. \citet{Darnstadt} prove the following version of Conjecture \ref{bendavid1}.
\begin{theorem} [{\citet[Theorem 1]{Darnstadt}}] \label{darnneg}
Let $H$ be a hypothesis class such that it contains the constant zero and constant one function. Then for every domain distribution $D$ and every $h \in H$,
\begin{enumerate}
\item if $H$ is finite then \begin{equation} \label{darneq1}
m(H,D_h,2 \epsilon,\delta) \leq O(\ln |H|) m^{\text{\normalfont SSL}}(H,D_h,\epsilon,\delta),
\end{equation}
\item if $H$ has finite VC-dimension then
\begin{equation} \label{darneq2}
m(H,D_h,2 \epsilon,\delta) \leq O(\vc(H)) \log(\frac{1}{\epsilon}) m^{\text{\normalfont SSL}}(H,D_h,\epsilon,\delta).
\end{equation}
\end{enumerate}
\end{theorem}

Note that this statement holds for all $D_h$, so in particular if we take the supremum over all $h \in H$ as in Conjecture \ref{bendavid1}. \citet{Golovnev} show that if the hypothesis class $H$ is given by the projections over $\{0,1\}^d$, there is a set of domain distributions such that any supervised algorithm needs $\Omega(\vc(H))$ as many samples as the semi-supervised counterpart, which has knowledge of the full domain distribution. So in particular Inequality \eqref{darneq2} is tight up to logarithmic factors. This actually shows that the constant improvement can be arbitrarily good, as we can increase the VC-dimension by increasing the dimension \citet[Proposition 4]{Golovnev}. The agnostic version of Theorem \ref{darnneg} is an open problem.

In the case of a hypothesis class with infinite VC-dimension, however, the conjecture ceases to hold, also for the slightly altered formulations. This is essentially the case because we can start with a class that has infinite VC-dimension, and thus cannot be learned by a supervised learner. A semi-supervised learner, however, can restrict this class in a way such that it has finite VC-dimension. We elaborate on this in the next section where we collect three different setups in which a semi-supervised learner can PAC-learn, while a supervised learner cannot.\footnote{In this context PAC-learnability means that $m(H,\epsilon,\delta)$ is finite for all $\epsilon, \delta>0$.}

\subsection{Impossibility When not Restricting the Possible Labeling Functions}

\citet{Golovnev} show that if the domain $\mathcal{X}$ is finite and we allow all deterministic labeling functions on it, no semi-supervised learner can improve over a consistent supervised learner in the realizable PAC-learning framework, not even by a constant. Consistent means here that the learner achieves $0$ training error. The supervised learner is, however, to be allowed twice as inaccurate and twice as unsure, which is respectively captured by the $2\epsilon$ and $2\delta$ in Inequality \eqref{golovneveq}.
\begin{theorem}[{\citet[Theorem 8]{Golovnev}}]
Let $\mathcal{X}$ be a finite domain, and let 
\[
H_{\text{\normalfont all}}=\{0,1\}^\mathcal{X}
\]
be the set of all deterministic binary labeling functions on $\mathcal{X}$. Let $A$ be any consistent supervised learner, $P$ a distribution over $\mathcal{X}$ and $\epsilon,\delta \in (0,1)$. Then
\begin{equation} \label{golovneveq}
m(A,H_{\text{\normalfont all}},P,2\epsilon,2\delta) \leq m^{\text{\normalfont SSL}}(H_{\text{\normalfont all}},P,\epsilon,\delta).
\end{equation}
\end{theorem}
While the more general Theorem \ref{darnneg} states that a semi-supervised can still be better by a constant depending on the hypothesis class, we find that in the previous setting one even loses this advantage.

A similar result can be found for the agnostic case.  Theorem 2 by \citet{gopfert} essentially states that Conjecture \ref{bendavid2} (the agnostic case), is true for the finite VC-dimension case, if there are no restrictions on the labeling function. The difference is that they consider in an in-expectation and not a high probability framework and there is a condition on the domain distribution $D$, while Conjecture \ref{bendavid2} is formulated to hold for \emph{all} distributions $D$. This condition is, however, very mild. The essential assumption of the theorem is that there are no restrictions on the labeling function.

The intuition for both of the previous results is the same: if we allow all labeling functions, there is no label information about the support of $\mathcal{X}$ that we did not observe yet. Finding the labels for this part is equally slow for supervised and semi-supervised learners. In the next section, we see that positive results are still possible and present hypothesis classes on which semi-supervised learners can be effective. Following the previous result, it is not surprising, however, that those classes are carefully chosen.

\section{On the Possibility of Semi-Supervised Learning}\label{effectivessl}

We consider three specific settings in which it can be shown that a semi-supervised learner can learn, while a SL cannot. We present the two works of \citet{Darnstadt} and \citet{Globerson}, these aim to answer Conjectures \ref{bendavid1} and \ref{bendavid2} covered in the previous subsection. They show that there is a hypothesis class $H^*$ and a collection of domain distributions $\mathcal{D}^*$ such that no supervised learner can learn $H^*$ under the distributions of $\mathcal{D}^*$. Given, however, any $P \in \mathcal{D}^*$, a semi-supervised learner that has access to a finite, but depending on $P$ arbitrarily large, amount of unlabeled data can learn $H^*$ with the same rate of convergence. As a third, we present the work of \citet{Niyogi} as we think it provides the most insightful example of how a shift from not learnable to learnable is possible when going from SL to SSL.

\subsection{Proving the Realizable Case with a Discrete Set}

\citet{Darnstadt} give the first example that shows that Conjecture \ref{bendavid1} does not hold in general. This is captured in the first theorem to follows. All other results, also covered in this section, basically follow a very similar line of reasoning.
\begin{theorem}[{\citet[Theorem 2]{Darnstadt}}] \label{Darn2}
There exists a hypothesis class $H^*$ and a family of domain distributions $\mathcal{D}^*$ such that
\begin{enumerate}
\item For every $D \in \mathcal{D}^*$, $$m^{\text{\normalfont SSL}}(H^*,D,\epsilon,\delta) \leq O(\frac{1}{\epsilon^2}+\frac{1}{\epsilon}\log(\frac{1}{\epsilon})).$$
\item For all $\epsilon < \frac{1}{2}$ and $\delta <1$,
$$m(H^*,\epsilon,\delta)= \sup\limits_{D \in \mathcal{D}^*} m(H^*,D,\epsilon,\delta)=\infty.$$
\end{enumerate}
\end{theorem}

In order for the semi-supervised learner to be able to PAC-learn for all $D \in \mathcal{D}^*$, it needs knowledge of the full distribution $D$. (Although for each fixed $D \in \mathcal{D}^*$, a finite amount of unlabeled data suffices.) Since the supervised learner can only collect labeled samples, it will never be able to achieve this knowledge with a finite number of samples and thus has an infinite sample complexity.  Results that show similar behavior using a hypothesis class that is loosely based on the manifold assumption are discussed in the next two sections. It is of interest to give the intuition for the example provided by \citet{Darnstadt}, as it uses the same trick that the other work in this section use as well.

\citet{Darnstadt} set the example up as follows. The domain $\mathcal{X}$ consists of all sequences
\[
x=(x_1,x_2,\ldots,x_{l})
\]
of arbitrary finite length and $x_i \in \{0,1\}$. The distributions $D \in \mathcal{D}^*$ on $\mathcal{X}$ are such that there is a sequence
\[
D(x_{\sigma(1)}=1) > D(x_{\sigma(2)}=1)>\dots,
\]
where $\sigma$ is a random permutation of the indices of $x$, and the distribution drops sufficiently quick in $\sigma(i)$.\footnote{Note that with $x_{\sigma(i)}=1$ we mean the subset $V \subset \mathcal{X}$ with

\[
V:=\{x=(x_1,x_2,\ldots,x_{l})\in \mathcal{X} \mid x_{\sigma(i)}=1\}.
\]}

The hypothesis class $H^*$ contains all hypotheses $h_i$ with $h_i(x)=x_i$ and the constant $0$ hypothesis. Note that although the class has infinite VC-dimension it still takes some effort to show that no supervised learner can learn it w.r.t to all distributions in $\mathcal{D}^*$. This is because the VC-dimension might not be infinite over $\mathcal{D}^*$. We want to sketch how the semi-supervised learner can learn it. After fixing a $D \in \mathcal{D}^*$ and $\epsilon, \delta >0$ we draw enough unlabeled samples to identify all positions $i \in \mathbb{N}$ such that $x_i$ is with a high probability $0$. For all those indices $i$ we can remove $h_i$ from $H^*$ as the constant $0$ hypothesis is good enough for predicting accurately. They then show that the remaining hypotheses in $H^*$ can be learned from finitely many samples.

The foregoing example, like those that follow, are essentially set up such that $H$ and $D$ have a certain link and in those cases knowledge about $D$ can actually give knowledge about $H$. Note, however, that knowledge about $D$ does not restrict the set of possible labeling functions from $H$. It is rather that $D$ helps to identify which hypotheses we can safely ignore.  Note, also that it is important that the admissible domain distributions are restricted. If $D^*$ would also include distributions that essentially put equal weight on all positions $i$, there would be no position $x_i$ which are with high probability $0$ and we thus could not remove the corresponding hypotheses.  
\begin{figure}[t!]
\centering
  \subfloat[]{%
    \includegraphics[width=0.35\textwidth,valign=t]{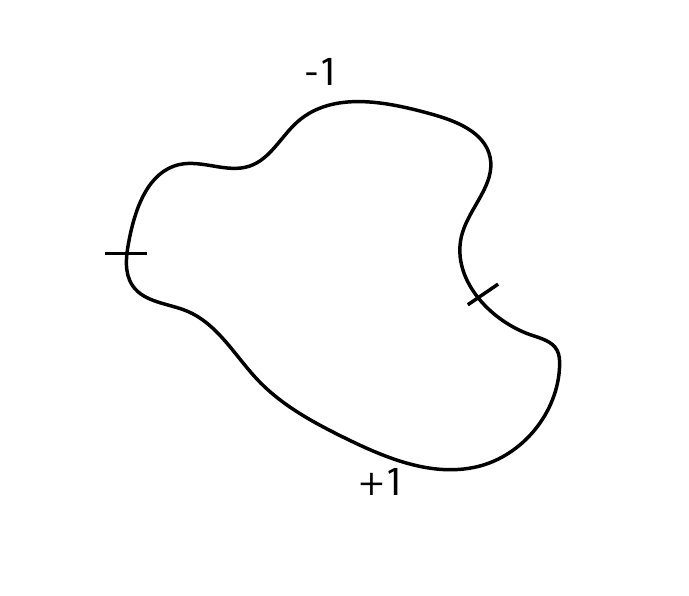}%
    \vphantom{\includegraphics[width=0.2\textwidth,valign=t]{circle1_small}}%
  } \quad
  \subfloat[]{%
    \includegraphics[width=0.35\textwidth,valign=t]{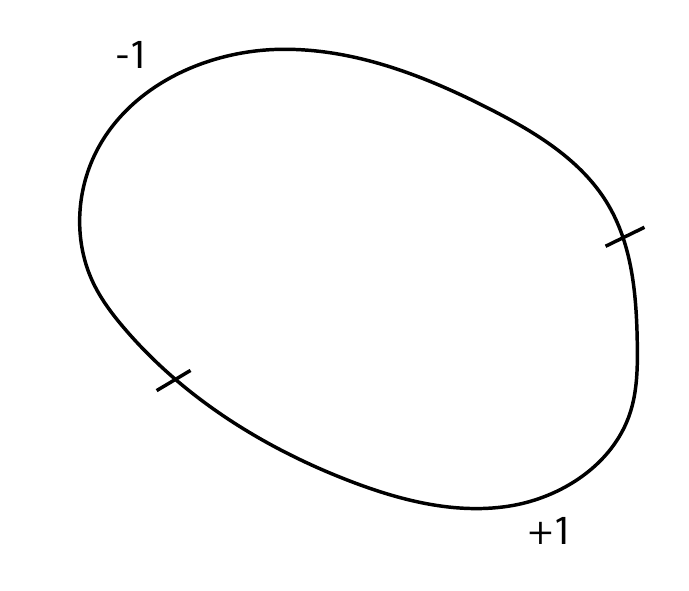}%
  }
  \caption{The shapes shown in (a) and (b) are two different embeddings of a circle in the euclidean plane. One half of the circle is labeled as $1$, while the other half is labeled as $-1$, while we assume that everything outside the circle is labeled as $1$.}
  \label{circles}			
\end{figure}
\subsection{Proving the Agnostic Case using Algebraic Varieties} \label{globerson}
\citet{Globerson} provide a different example where the set of admissable distributions are given by specific manifolds. So they use the second, alternative, version of the manifold assumption introduced in Section \ref{ssassumptions}. Using this example one can also show that Conjecture \ref{bendavid2}, so the impossibility conjecture for the agnostic case, is not true in general. The theorem is very similar to the one from \citet{Darnstadt} that was discussed above. The main difference is in the construction of the hypothesis set and the set of distributions.
\begin{theorem}[{\citet[Theorem 5]{Globerson}}] \label{manifoldthm}
There exists a hypothesis class $H_{\normalfont \text{alg}}$ and a set of distributions $\mathcal{D}_{\normalfont \text{alg}}$ such that
\begin{enumerate}
\item for every $D \in \mathcal{D}_{\normalfont \text{alg}}$,
\begin{equation} \label{manifold1}
m^{\text{\normalfont SSL}}(H_{\normalfont \text{alg}},D,\epsilon,\delta)<\frac{2}{\epsilon}\log \frac{2}{\delta};
\end{equation}
\item the supervised sample complexity is infinite, i.e.,
\begin{equation} \label{manifold2}
 \sup\limits_{D \in \mathcal{D}_{\normalfont \text{alg}}} m(H_{\normalfont \text{alg}},D,\epsilon,\delta)=\infty.
\end{equation}
\end{enumerate}
\end{theorem}

The hypothesis class $H_{\normalfont \text{alg}}$ consists of all hypotheses that have class label $1$ on an algebraic set and $0$ outside of that set. This algebraic set can essentially be considered a form of manifold.  This hypotheses class is still a very expressive and has infinite VC-dimension. If, however, we restrict the set of admissible domain distributions $\mathcal{D}_{\normalfont \text{alg}}$ to be particular types of algebraic sets, a semi-supervised learner with knowledge of $D \in \mathcal{D}_{\normalfont \text{alg}}$ can learn efficiently. We can think of $\mathcal{D}_{\normalfont \text{alg}}$ as the set of distributions that have support on a finite combination of distinguishable algebraic sets $V_1,\ldots,V_k$. Once we know that the distribution has support on $V_1,\ldots,V_k$, we only have to figure out which of those algebraic sets have label $1$ and which have label $0$.  A semi-supervised learner can thus reduce the class $H_{\normalfont \text{alg}}$ by only considering the hypotheses that have class label $1$ on combinations from $V_1,\ldots,V_k$. Since the set of all possible combinations is finite, a semi-supervised learner can learn them with a sample complexity bounded by Inequality \eqref{manifold1}.  Note that although the true labeling function does not have to be part of this restricted set, one can show that it is anyway always optimal to predict with a hypothesis from it. The argument for it is similar to the explanation of the agnostic case, which we covered in the following.

When the true target function is not in $H_{\normalfont \text{alg}}$, the above extension might appear problematic at first, because the semi-supervised algorithm restricts the hypothesis set $H_{\normalfont \text{alg}}$. To guarantee PAC-learnability we need to know that the best predictor from the $H_{\normalfont \text{alg}}$ is still in this restricted set. But this is indeed the case, because the set of domain distributions $\mathcal{D}_{\normalfont \text{alg}}$ was exactly created for that to hold. To show this, assume that the distribution is supported on an irreducible algebraic set $V_0$. Our semi-supervised learner can now choose to label it completely $1$ or $0$, while both options might lead to non-zero error. But labeling it completely as either $1$ or $0$ is already ideal, as using any other algebraic set $V_1 \in H_{\normalfont \text{alg}}$ will lead to one of those two labelings. This is because, by construction, $V_1$ is either equal to $V_0$ (which leads to label everything as $1$) or has an intersection of zero mass (which leads to labeling almost everything as $0$).

Interestingly, the above mentioned findings seems to contradict the results from Subsection \ref{onlymanifold}, as \citet{Lafferty} show that a supervised learner can also adapt to the underlying manifold. This discrepancy is not easy to analyze, as \cite{Lafferty} work in the regression setting, while \citet{Globerson} analyse classification. The intuition, however, is that \citet{Globerson} present the supervised learner with an impossible, meaning not PAC-learnable, task. \citet{Lafferty} on the other hand restrict the target functions to be smooth and thus the supervised learner is presented with a sufficiently easy problem.

\subsection{Using the Manifold Assumption to Make A Class Learnable}\label{niyogi}

\begin{figure}[t!]
\centering
\subfloat[Assume we are given $7$ points that are labeled as depicted above.]{%
  \includegraphics[width=0.35\textwidth]{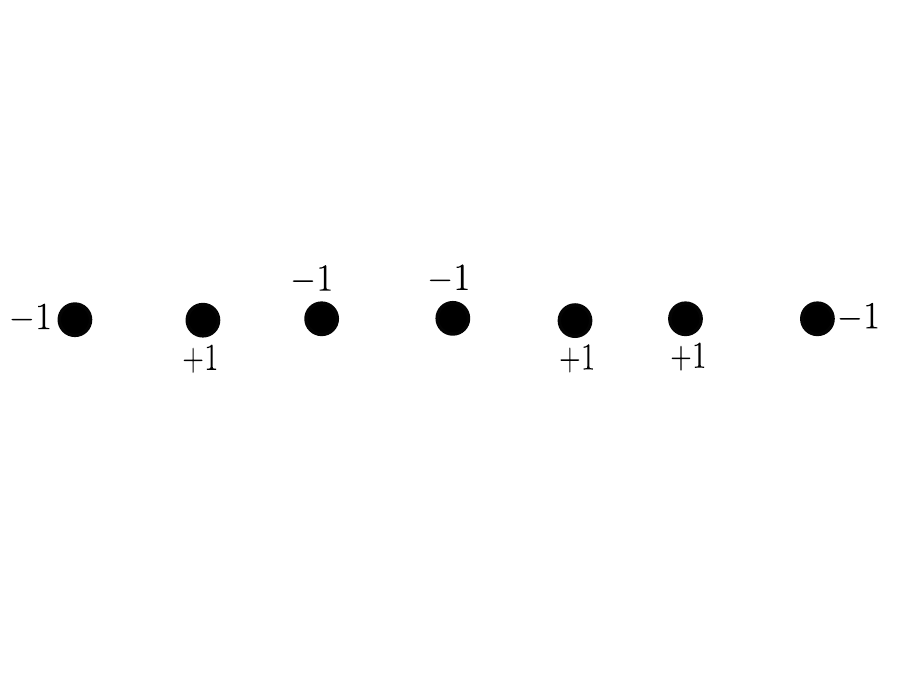}%
  }
  \hphantom{\includegraphics[width=0.2\textwidth,valign=t]{Infinite_VC_Circles_small}}%
    \vphantom{\includegraphics[width=0.15\textwidth,valign=t]{Infinite_VC_Circles_small}}%
\subfloat[The circle above labels the points correctly. The upper half assigns points the label $-1$, while the lower half labels points as $+1$.]{%
  \includegraphics[width=0.35\textwidth]{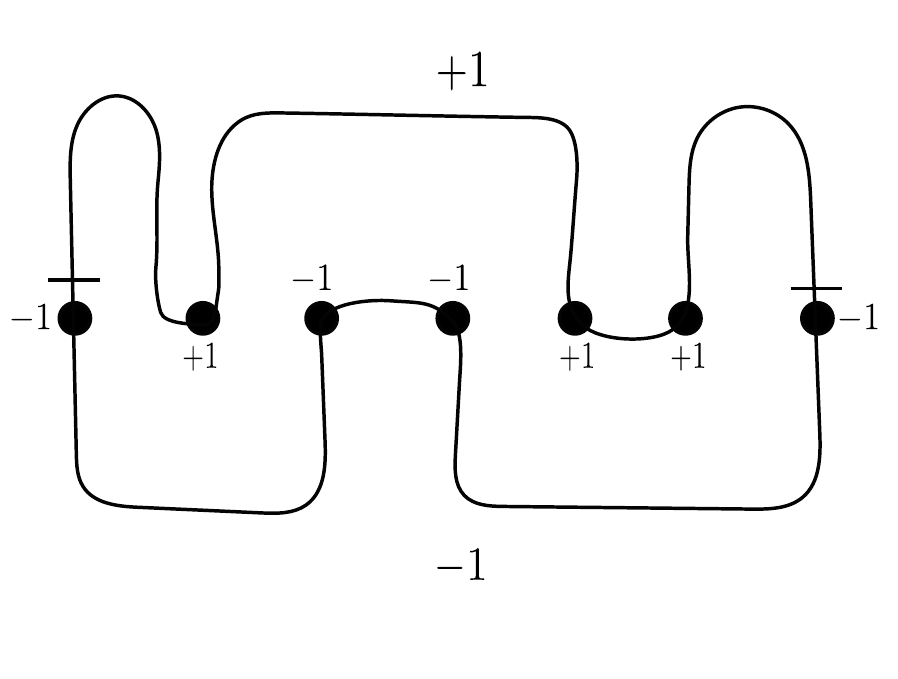}%
  }
\caption{A schematic proof why the hypothesis set $H_c$ has an infinite VC dimension. Given the points in (a) we can label them correctly with the circle given in (b).}
\label{circlesinfinitevc}
\end{figure}

\citet{Niyogi} provides another setup in which a semi-supervised learner can effectively learn, while a supervised learner cannot. The motivation for this, however, was independent of \citep{BenDavid} and meant as a general theoretical analysis of the manifold learning framework as introduced in \citep{Belkin}.  Also, their results are in-expectation, while the previous papers give PAC bounds, i.e., they hold with high probability.  This work is based on the manifold assumption, so a given domain distribution does limit the possible labeling functions, and thus is not a counterexample to \ref{bendavid1}. We believe, however, that it is the most intuitive setting to understand why a supervised learner cannot learn, while a semi-supervised learner can.

The paper presents the results in an in-expectation framework, we alter the setup slightly and present it in the PAC learning framework. We believe that also in this setting it is perfectly possible to convey the core ideas from \citep{Niyogi} and it allows us to draw better connections to the previously discussed papers.

The example is build up as follows. First it is assumed that the admissible domain distributions are given by the class of distributions $\mathcal{P}_c$ that have support on embeddings of a circle in the Euclidean plane (see Figure \ref{circles}). The hypothesis class $H_c$ consists of all possible binary labelings of half circles, while everything outside the circle is labeled as $1$\footnote{The labeling outside of the circle is a formality to ensure that the supervised learner makes predictions for the whole circle, as the learner does not a priori know in which part of the space the circle is embedded.}. The semi-supervised learner that knows the specific embedding of the circle only needs to find two thresholds on the given circle.  This is a hypothesis class with a VC-dimension of 2, which implies that the semi-supervised learner can learn efficiently. In Figure \ref{circlesinfinitevc}, we illustrate in a schematic way why $H_c$ has an infinite VC dimension and thus cannot be learned by any supervised learner.

\chapter{Learning without Assumptions}\label{Noassumptions}

As we have seen in the previous chapter, it can be difficult to use unlabeled data without any additional assumptions. In fact, we saw that in various of these situations one can show that unlabeled data cannot help at all. As already mentioned in the introduction of Chapter \ref{Impossibility}, this impossibility sometimes stems from the fact that we only consider improvements of the estimate of the conditional probability $P(Y|X)$.  The works we present in this chapter looks at the complete risk  $\mathbb{E}_{X,Y} \left[ l(h(X),Y)\right]$, a quantity which is always influenced by the marginal distribution $P(X)$. Still, no additional assumptions about the distribution $P$ are considered and the theoretical guarantees are accordingly weak.

We first present the work by \citet{Sokolovska}, who use the unlabeled data to reweigh the labeled points and show improvements in terms of asymptotic efficiency. Interestingly, one needs that the model is misspecified to show this result. Second, we present the work by \citet{Kaariainen}, who uses the unlabeled data to determine the center of the version space. The best possible improvements in the learning rate as reported in that work are bounded by a factor of $2$. Finally, we present the work of \citet{Leskes}, who uses unlabeled data to combine different hypothesis spaces and shows that the learning rates depend on the highest Rademacher complexity\footnote{The Rademacher complexity is a notion for the capacity of a hypothesis class, similar to the VC dimension.} amongst those hypothesis spaces.

\section{Reweighing the Labeled Data by the True Marginal}\label{sokolovska}

\citet{Sokolovska} proposed a semi-supervised learner that uses knowledge of the marginal distribution $P(X)$ in a reweighing scheme. In order to avoid difficulties for the theoretical analysis, they restrict the feature space $\mathcal{X}$ to contain only finitely many points and assume that the semi-supervised learner has access to the full marginal distribution $P(X)$.\footnote{This line of work is continued by \citet{Kawakita}, where it is extended to non-discrete features spaces.} They consider models that directly estimate class probabilities $p(y|x, \theta)$, while they measure performance by the negative log-likelihood
\[
l(x,y|\theta)=-\ln p(y|x,\theta).
\]
They then analyze asymptotic behavior, in particular the asymptotic variance of the model estimation. In the end, two models are compared, the classical maximum log-likelihood estimate based on the labeled data only:
\begin{equation}
\theta^{\text{\normalfont SL}}=\arg \min\limits_{\theta \in \Theta} \sum_{(x,y) \in S_n} l(x,y|\theta)
\end{equation}
and a semi-supervised learner that also takes the marginal $P(x)$ into account:
\begin{equation} \label{reweighted}
\theta^{\text{\normalfont SSL}}=\arg \min\limits_{\theta \in \Theta} \sum_{(x,y) \in S_n}  \frac{P(x)}{\sum_{z \in X_n } I_{\{x=z \}}} l(x,y|\theta).
\end{equation}
Note that the semi-supervised learner weighs each feature with the true, instead of the empirical distribution.

We now first state the results concerning $\theta^{\text{\normalfont SSL}}$ after which we discuss them.
\begin{theorem} [{\citet[Theorem 1]{Sokolovska}}]
Let
\[
\theta^{*} \in \arg \min\limits_{\theta \in \Theta} \mathbb{E}[  l(x,y|\theta)]
\]
and define the following matrices
\begin{align}
& H(\theta^*)=\mathbb{E}_{X} \left[ \mathbb{V}_{Y|X}[ \nabla_{\theta} l(X,Y|\theta) | X ] \right] \\
& I(\theta^*)=\mathbb{E}_{X,Y} \left[  \nabla_{\theta} l(X,Y|\theta)  \nabla_{\theta}^T l(X,Y|\theta)   \right] \\
& J(\theta^*)=\mathbb{E}_{X,Y} \left[  \nabla_{\theta}^T \nabla_{\theta} l(X,Y|\theta)      \right],
\end{align}
where $\mathbb{V}_{Y|X}$ is the variance over the conditional random variable $Y|X$. Then $\theta^{\text{\normalfont SL}}$ and $\theta^{\text{\normalfont SSL}}$ are consistent and asymptotically normal estimators of $\theta^*$ with
\begin{equation}
\sqrt{n}(\theta^{\text{\normalfont SL}}-\theta^*) \rightarrow \mathcal{N}(0,J^{-1}(\theta^*)I(\theta^*) J^{-1}(\theta^*))
\end{equation}
\begin{equation}
\sqrt{n}(\theta^{\text{\normalfont SSL}}-\theta^*) \rightarrow \mathcal{N}(0,J^{-1}(\theta^*)H(\theta^*) J^{-1}(\theta^*))
\end{equation}
and $\theta^{\text{\normalfont SSL}}$ is asymptotically efficient, meaning that it achieves asymptotically the smallest variance of any unbiased estimator.
\end{theorem}

Asking now when $\theta^{\text{\normalfont SSL}}$ asymptotically dominates $\theta^{\text{\normalfont SL}}$, we get the, at a first glance, somewhat surprising answer that this actually happens when the model is misspecified. We know, however, that it cannot happen if the model is well-specified, as in that case (along with some other regularity conditions) the MLE $\theta^{\text{\normalfont SL}}$ is already asymptotically efficient itself. Specifically, we have in this case that
\[
H(\theta^*)=J(\theta^*)=I(\theta^*)
\]
and we recover the classical result that the MLE is asymptotically normal with a variance of the inverse fisher information matrix $I(\theta^*)$.

The paper examines, based on the logistic regression model, when the difference between $I(\theta^*)$ and $H(\theta^*)$ is particularly big and shows that this is the case the more $P(Y|X)$ is bounded away from $1/2$, so in particular when the Bayes error is small. Such requirement on $P(Y|X)$ is very similar to the \emph{Tsybakov-margin condition} \citep{tsybakov}, which is used in statistical learning to come to fast learning rates. In Sections \ref{Exponential} and \ref{cluster1}, similar assumptions are made to show that some semi-supervised learners can converge exponentially fast to the Bayes error.

\section{Picking the Center of the Version Space} \label{kari}

\citet{Kaariainen} introduces a method for bounding the risk by using unlabeled data to collect information about the agreement of two classifiers. A semi-supervised estimator is then derived as the hypothesis that minimizes this bound. Unfortunately, the idea only really works in the realizable case. Although we do not get a new algorithm for the agnostic case, the paper still presents new bounds for supervised methods that make use of the unlabeled data.

\subsection{Realizable Case}
The idea for the realizable case is to consider the version space, i.e., the space that contains all hypotheses that have no training error. The unlabeled data gives rise to a pseudo-metric on this space by measuring the disagreement of its hypotheses on this data. We are then going to take the hypothesis that has the lowest worst-case disagreement to all other hypothesis. One of must be the true one as we assume realizability. Let us now make this more precise.

Given two hypotheses $f,g \in H$ we define the disagreement pseudo-metric $d(f,g)$ as
\begin{equation}
d(f,g)=P(f(X) \neq g(X)).
\end{equation}
This metric is specifically useful in the semi-supervised case since is does not depend on labels. We can approximate it with the empirical version by
\begin{equation} \label{versionspaceestimator}
\hat{d}(f,g)=\frac{1}{m}\sum_{i=n}^{n+m} I_{\{f(x_i)=g(x_i)\}}.
\end{equation}
The version space is defined as
\[
H_0=\{h \in H \  | \ \hat{R}(h)=0 \}.
\]
If $h_0$ is the true hypothesis, then we know that $h_0 \in H_0$ and one can show that $R(h)=d(h,h_0)$ for all $h \in H$. This, in  turn, gets us to the following bound.
\begin{equation} \label{versionspace}
\begin{split}
R(h) =d(h,h_0)=\hat{d}(h,h_0)+(\hat{d}-d)(h,h_0) & \\ \leq \sup\limits_{g \in H_0} \hat{d}(h,g)+  \sup\limits_{g,g' \in H_0} (\hat{d}-d)(g,g'). &
\end{split}
\end{equation}
As Inequality \eqref{versionspace} bounds the true risk of a hypothesis $h$, we try to minimize this risk by choosing the hypothesis that minimizes the right-hand side of Inequality \eqref{versionspace}.
More specifically, we choose the semi-supervised estimator to be the so-called \emph{empirical center of the version space}, which means we take
$$h^{\text{\normalfont SSL}}=\arg \inf\limits_{h \in H_0} \sup\limits_{g \in H_0} \hat{d}(h,g).$$

With this we can of course only control the first term on the right-hand side of Inequality \eqref{versionspace}. In a standard way, we can bound the second term with concentration inequalities derived from a Rademacher complexity for the space
\[
\mathcal{G}=\{ x \mapsto I_{\{f(x)=g(x)\}} \ | \ f,g \in H_0 \}.
\]
Ultimately, this leads us to the result that with probability at least $1-\delta$ \citep[Theorem 3]{Kaariainen}
\begin{equation} \label{disagreementresult}
R(h^{\text{\normalfont SSL}}) \leq \inf\limits_{h \in H_0} \sup\limits_{g \in H_0} \hat{d}(h,g) + \empRad(\mathcal{G})+\frac{3}{\sqrt{2}} \sqrt{\frac{\ln\frac{2}{\delta}}{m}}.
\end{equation}

Note the two terms on the right-hand side of Inequality \eqref{disagreementresult} go to $0$ for increasing $m$ and that, in this case, we also have that $\hat{d}(f,g) \rightarrow d(g,g)$. So ignoring for a minute that we only have finitely many unlabeled data points, we can compare the semi-supervised learner to purely supervised solutions. Note that in the realizable case a purely supervised method would also choose a hypothesis in $H_0$. As the supervised learner $h^{\text{SL}}$ has no additional information, we can always find a target hypothesis $h^*$ such that
\[
R(h^{\text{SL}})=\sup_{g \in H_0}d(h^{\text{SL}},g)=d(h^{\text{SL}},h^*).
\]
So the best bound for any supervised learner $h^{SL}$ is given by
\[
R(h^{SL}) \leq \sup_{g \in H_0}d(f,g).
\]
The SSL bound \eqref{disagreementresult}, on the other hand, allows us to come to the following bound:
\[
R(h^{\text{SSL}}) \leq \inf_{h \in H_0} \sup_{g \in H_0} d(h,g),
\]
which holds at least for $m$ going to infinity.

From a geometrical viewpoint, $\sup_{g \in H_0}d(h^{\text{SL}},g)$ is the diameter of $H_0$, while, $\inf_{h \in H_0} \sup_{g \in H_0} d(h,g)$ is the radius. As the difference between the radius and the diameter, with respect to $d$, is at most 2, we find that the differences in the SSL and SL risk bounds is at most a constant factor of 2.

\subsection{Bounds for the General Case}

In the general, agnostic case, we do not assume that the target hypothesis is part of our hypothesis class. To still make use of the considered disagreement pseudo-metric, the author proposes the following general recipe for bounds in that case

The starting point is the observation that bounds for randomized classifiers are generally tighter compared to their deterministic counterparts \citep{McAllester3, Langford}. The idea is now to use such a randomized classifier $f_{\text{rand}}$ as a kind of anchor.  This anchor takes on a role similar to the target hypothesis in the realizable case. To get a bound for a classifier $f$, we can use the bound for the randomized classifier together with a slack term that includes $\hat{d}(f_{\text{rand}},f)$. Depending on which kind of randomized classifier we take, we obtain different bounds. This includes for example PAC-Bayesian bounds as well as bounds based on cross-validation and bagging methods.  \citet{Kaariainen} additionally derives an explicit cross-validation bound, where the randomized classifier is given by a uniform distribution over the classifiers obtained in the multiple cross-validation rounds.

\section{Combining Multiple Hypothesis Spaces}

\citet{Leskes} presents another semi-supervised scheme that relies on a measure of the classification agreement between hypotheses on unlabeled data. The idea here is to use an ensemble scheme, which means we start with $L \in \mathbb{N}$ different hypothesis classes $H^1,\ldots,H^L$. We want to find the best fitting hypothesis over all $L$ hypothesis classes $H^1,\ldots,H^L$. As that would generally lead to an overly increased complexity, the paper reduces the set of possible hypotheses by only considering those that agree sufficiently on the unlabeled data. In this context sufficiently means that we switch to a new hypothesis class $H_v$ for a $v>0$ that is defined as
$$
H_v=\{(h^1,\ldots,h^L) \in H^1 \times \ldots \times H^L \ | \ V(h^1,\ldots,h^L) \leq v \},
$$
where
$$
V(h^1,\ldots,h^L):=\mathbb{E}_X [\frac{1}{L}\sum\limits_{i} h^i(X)^2-(\frac{1}{L}\sum\limits_{i} h^i(X))^2].
$$
The term $V(h^1,\ldots,h^L)$ essentially measures the variance of disagreement within $L$ different hypotheses and is approximated with the unlabeled data. The hypothesis class $H_v$ only keeps those collections of hypotheses that have a sufficiently small variance of disagreement. The paper then presents a generalization bound that holds for all $h^l$ with $1 \leq l \leq L$ simultaneously, where the bound depends on the maximum Rademacher complexity of the $L$ base hypothesis classes $H^1,\ldots,H^L$.

\chapter{Learning under Weak Assumptions} \label{weakassumptions}

In the previous two chapters, we investigated what is possible for semi-supervised learners when we do not have any additional assumptions. Here we investigate what a semi-supervised learner can achieve under what we call \emph{weak} assumptions. With weak assumptions we mean those that cannot essentially change the learning rate of $O(\frac{1}{\sqrt{n}})$, but rather gives improvements by a constant which can depend on the hypothesis class. In Chapter \ref{strongassumptions}, we investigate what we have to assume to escape the $\frac{1}{\sqrt{n}}$ regime, and also elaborate on the differences. 

Here, we first cover the work of \citet{Balcan}, as it provides a rather general framework that allows us to analyze the learning guarantees for various semi-supervised learners. This initial paper shows that semi-supervised learners that fall in this framework learn by a constant faster then supervised learners, where the constant depends on the hypothesis class and the semi-supervised learner we use.  We then cover in more detail the idea of co-training. Co-training can also be studied within the setting \citet{Balcan} consider, but we present some additional details of interest not fully captured by this framework. In particular, we present the work by \citet{Sridharan}, who formulate the assumption of co-training in an information theoretical framework, which allows one to precisely quantify the bias-variance trade-off.

\section{A General Framework to Encode Weak Assumptions} \label{sectionbalcan}

The work done by \citet{Balcan} offers an elegant way to formalize different assumptions in a general framework. Many existing methods can be cast in this framework: transductive support vector machines \citep{Joachims,Boyd}, multi-view assumptions \citep{Blum,Leskes,Sridharan}, and transductive graph-based methods \citep{Blum2} are just some examples. The idea is to introduce a function $\chi$ that measures the compatibility between a hypothesis $h$ and the marginal distribution $P(X)$. Compatibility can mean many different things in this context. As a simple example, we can deem a hypothesis $h$ compatible with a marginal distribution $P(X)$, if its decision boundary goes through low density regions.

We want to define the function $\chi$ for each point in the feature space, so that we can compute averages of $\chi$ based on finite unlabeled samples. So one sets 
\begin{equation} \label{compatibility}
\chi: H \times \mathcal{X} \to [0,1].
\end{equation}
The compatibility measure $\chi$ then gives rise to the function
\begin{equation} \label{Runl}
R_{\text{\normalfont unl}}(h):=1 -\mathbb{E}_{X \sim P(X)} [\chi(h,X)],
\end{equation}
which will be referred to as the \emph{unsupervised loss}. Our aims is to optimize it in addition to the loss measured on the labeled sample.

The paper states several theorems, but we focus on only one of them. The other results have, however, the same flavor as the one presented here. The differences are mostly in the realizability assumptions, i.e., regarding the unsupervised and the supervised error, and the bounding technique. \citet{Balcan} present bounds derived from uniform convergence as well as bounds based on covering numbers. The following theorem considers the double agnostic case in which neither the labeled nor the unlabeled loss have to be zero.
\begin{theorem}[{\citet[Theorem 10]{Balcan}}] \label{balcan1}
Let
\[
h^*_t=\arg \min\limits_{h \in H}[R(h) | R_{\text{\normalfont unl}}(h) \leq t].
\]
Then, given an unlabeled sample size of at least $$
\mathcal{O}\left( \frac{\max[VC(H),VC(\chi(H))}{\epsilon_2} \ln \frac{1}{\epsilon_2} +\frac{1}{\epsilon_2^2} \ln \frac{1}{\delta} \right)
$$
we have that
\begin{equation}\label{balcanSampleComplex}
m(h^{\text{\normalfont SSL}},H,\epsilon,\delta) \leq \frac{32}{\epsilon^2} \left[ VC(H(t+2\epsilon_2))+\ln \frac{2}{\delta} \right],
\end{equation}
where $h^{\text{\normalfont SSL}}$ is the hypothesis that minimizes $\hat{R}(h^{\text{\normalfont SSL}})$ subject to
\[
\hat{R}_{\text{\normalfont unl}}(h^{\text{\normalfont SSL}}) \leq t +\epsilon
\]
and
\[
H(t):=\{ h \in H \ | \ R_{\text{\normalfont unl}}(h) \leq  t\}.
\]
Here $\hat{R}$ is the empirical risk measured with the sample $S_n$ and $\hat{R}_{\text{\normalfont unl}}$ is the empirical unlabeled risk measured on the sample $U_m$.
\end{theorem}
It is important to note here that the original paper uses a measure of complexity different from the term $VC(H(t+2\epsilon_2))$. To allow for an easier comparison to other results in this review (and to avoid additional notation), we express the above theorem in terms of the standard VC-dimension. The original use a complexity notion that can be found in \citep{Vapnik1998} as the (exponentiated) annealed entropy.  The advantage of the latter measure is that it is distribution dependent.

Let us now indeed briefly compare Theorem \ref{balcan1} to results from the previous chapter.  In particular, let us consider Conjecture \ref{bendavid1} and the answers to this as found in Theorems \ref{Darn2} and \ref{manifoldthm}. We know that in the purely supervised case, we can achieve a similar sample complexity as in Equation \eqref{balcanSampleComplex} by replacing $VC(H(t+2\epsilon_2))$ with $VC(H)$. As we know that the complexity given by Equation \eqref{balcanSampleComplex} is tight up to constants \citep[see also][Chapter 6]{Shalev}, we know that the sample complexity between a purely supervised learner and the semi-supervised learner as defined in this paper cannot differ by more than $\mathcal{O}\left(\frac{VC(H)}{VC(H(t+2 \epsilon_2)}\right)$. So the gap in the learning rates is indeed given by a constant that only depends on the hypothesis class as postulated by Conjecture \ref{bendavid2}. This constant can, however, be infinite if $VC(H)$ is infinite but $VC(H(t+2 \epsilon_2))$ is finite. This is exactly the type of example that refutes the conjecture and that we reviewed in Section \ref{effectivessl}.

Theorem \ref{balcan1} quantifies, to some degree, the fundamental bias-variance trade-off in SSL when we rely on additional assumptions. Employing a semi-supervised compatibility function, we reduce the variance of the training procedure as we effectively restrict the original hypothesis space $H$.  If, however, the compatibility function does not match the underlying problem, we at the same time bias the procedure away from good solutions.

\section{Assuming that the Feature Space can be Split}\label{multiview}

In multi-view learning, incidentally also referred to as co-regularization or co-training, one assumes that the feature space $\mathcal{X}$ can be decomposed as 
\[
\mathcal{X}=\mathcal{X}^1 \times \mathcal{X}^2
\]
and each partial feature space $\mathcal{X}^1,\mathcal{X}^2$ is, in principle, enough to learn. In the early work on co-training, \citet{Blum} use the idea in a web page classification set. One part of the features, say $\mathcal{X}^1$, is given by the text on the web page itself, while the other one, $\mathcal{X}^2$, is given by the anchor text of hyperlinks pointing to the web page. The idea is that if both partial features spaces have sufficient information about the correct label, we would expect that a correct classifier predicts the same label given any of the two partial features. We can thus discard classifiers that disagree on the two views, and this disagreement can be measured with unlabeled data.

There are multiple theoretical results that pertain to this approach. It can, for example, be analyzed in the framework of the previous chapter.  Alternatively, \citet{Rosenberg} and \citet{Farquhar} analyze a Rademacher complexity term under the multi-view assumption, while \citet{Sindhwani} define a kernel that directly includes the assumption as a regularization term, and thus find a RKHS where co-regularization automatically applies.  Here we detail the work of \cite{Sridharan} as it ties in best with the other results we present. In addition, their information theoretic framework allows us to also analyze the penalty one suffers if the assumption is not exactly true.

As above, we split the random variable $X$, which takes values in $\mathcal{X}$, into two: $X=(X^1,X^2)$. Now, the multi-view assumption from \citep{Sridharan} can be formalized as follows: let $I(A;B|C)$ be the mutual information between random variables $A$ and $B$, conditioned on knowing already the random variable $C$. Then there exists an $\epsilon_{\text{info}}$ such that
\begin{equation}
I(Y;X^2|X^1) \leq \epsilon_{\text{info}}
\end{equation}
and
\begin{equation}
I(Y;X^1|X^2) \leq \epsilon_{\text{info}}.
\end{equation}
Intuitively, this states that once we know one of the features, the other feature do not tell us much more about $Y$.  Comparing this to co-training, we can see it as a relaxation: in co-training one assumes that each view is already sufficient to fully learn, which would corresponds to an $\epsilon_{\text{info}}$ that equals $0$.  If, however,  $\epsilon_{\text{info}}>0$, we cannot learn perfectly from one view.

We now assume, subsequently, that we have for each view $X^1$ and $X^2$ a corresponding hypothesis set $H^1$ and $H^2$. We carry out predictions with \emph{pairs} of hypotheses
\[
(f_1,f_2) \in H^1 \times H^2.
\]
The paper uses the notion of compatibility functions, as generally defined through Equation \eqref{compatibility}. In particular, they define the compatibility function
\[
\chi: H:=H^1 \times H^2 \to [0,1]
\]
as
\[
\chi(h^1,h^2,x):=d(f_1(x^1),f_2(x^2)),
\]
where
\[
d:\mathcal{Y} \times \mathcal{Y} \to [0,1]
\]
is a specific pseudo-distance measure that fulfills a relaxed triangle inequality and $x=(x^1,x^2)$ is a sample. In essence, the distance $d$ measures how much $f_1$ and $f_2$ agree on a sample $x$. For a given threshold $t \in \mathbb{R}$ we then find the best \emph{pair} of hypotheses with the constrained empirical risk minimization problem
\begin{equation}  \label{multivewsolution}
\min\limits_{(h^1,h^2) \in H} \sum_{i=1}^{n}l(h^1(x^1_i),y_i)+l(h^2(x^2_i),y_i)
\end{equation}
subject to $\hat{R}_{\text{\normalfont unl}}(h^1,h^2) \leq t$.

The main theorem, which gives guarantees on the solution found by the procedure above, needs the following notation. Let $\beta_*$, $\beta_*^1$ and $\beta_*^2$ be the Bayes error, measured with the loss $l$, when learning from $X^1 \times X^2$, $X^1$ and $X^2$ respectively. We also set
\[
\epsilon_{\text{bayes}}=\max \{R(f^1_*)-\beta^1_*, R(f^2_*)-\beta^2_* \},
\]
where $f^i_*$ is the best predictor from $H^i$. Finally, recalling the definition of $R_{\text{\normalfont unl}}(h)$ from Equation \eqref{Runl}, we define
\[
\hat{H}(t)=\{ (h^1,h^2) \in H \ | \ \hat{R}_{\text{\normalfont unl}}(h^1,h^2) \leq t \}.
\]

\begin{theorem}[{\citet[Theorem 2]{Sridharan}}]\footnote{We should note that the theorem actually relies on some additional regularity conditions on $\chi=d$ and the loss $l$, which we have not made explicit so as to not distract from the main point we want to highlight.}
Given that the loss $l$ is bounded by $1$, there exists a $t \in \mathbb{R}$ (depending among others on $\epsilon_{\text{\normalfont info}}$, $\epsilon_{\text{\normalfont bayes}}$, and $m$) such that given at least $m(\hat{H}(t),\epsilon,\delta)$ labeled samples, with probability $1-\delta$
\begin{equation} \label{multiviewbound}
\frac{R(\hat{h}^1)+R(\hat{h}^2)}{2} \leq \beta_*+\epsilon+\epsilon_{\text{\normalfont bayes}}+\sqrt{\epsilon_{\text{\normalfont info}}}.
\end{equation}
\end{theorem}
Here we can see now that the information theoretic assumption allows us to describe the bias introduced when switching from the full hypothesis set $H$ to the restricted one $\hat{H}(t)$. In fact, this bias is given explicitly by $\sqrt{\epsilon_{\text{info}}}$.

\chapter{Learning under Strong Assumptions} \label{strongassumptions}

In the previous chapter, we analyzed assumptions that only could give us improvements in terms of a multiplicative constant. These did not allow us to come to semi-supervised learners that improve beyond the general learning rate of $\frac{1}{\sqrt{n}}$. Here, we analyze assumptions that enable us to escape this regime, even leading to exponentially fast convergence in some cases. We want to iterate a remark from the outline: there is not necessarily a principled qualitative difference in weak and strong assumptions, but rather a subtle quantitative difference. Both might make assumptions about the data distribution, but strong assumptions are just that, stronger. To gain for example exponential rates with a cluster assumption, as we show in Section \ref{cluster1}, we need to assume that the posterior distribution $P(Y \mid X)$ is bounded away from $\frac{1}{2}$. If we would, however, encode some type of cluster assumption in the framework we introduced in Section \ref{sectionbalcan}, which we considered a weak assumption, we could not get exponential rates. We could for example encode the manifold assumption, but we would only get constant improvements, and this was actually shown by \cite{Mey}. The difference is that in the framework of Section \ref{sectionbalcan} one can't enforce an assumption about $P(Y \mid X)$. We could enforce that the best solution is one that separates two clusters, but deciding which cluster belongs to which class can still be in the order of $\frac{1}{\sqrt{n}}$, for a class probability $P(Y \mid X)$ approaching $\frac{1}{2}$.

Before reviewing the actual works that present such results, we briefly describe a simple and basic example setting that, in essence, illustrates the main idea behind such possibilities for improvement.  In this setting, we come to a clustering based on all of the data provided and assume that the clustering is correct.  By the latter, we mean to indicate that each cluster corresponds to one class only.  Under this assumption, we only need enough labeled data to identify which cluster belongs to which class. The work we present in the current chapter extends this idea in various ways and answers the following questions. What if we have class overlap?  What if there is noise in the clusters, such that the clustering may not be completely correct?  How do we can beyond the classification setting and deal with regression?

\section{Assuming the Model is Identifiable} \label{Exponential}

One of the classic analyses in semi-supervised learning that deals with a topic closely related to the sample complexity, was carried out by \citet{Castelli}. As it turns out, the setting is quite restrictive but can, as such, give exponentially fast convergence to the Bayes risk in the number of labeled samples $n$. In a sense, the outcome is very strong, considering that the results covered in the previous chapters were essentially unable to improve upon the standard convergence rate of $\frac{1}{\sqrt{n}}$.  See for instance Inequality \eqref{balcanSampleComplex} after solving for $\epsilon$.  

The first key assumption to actually obtain these results lies in the data generation process. To start with, the label is supposed to be drawn with $P(y=1)=\eta$ and $P(y=0)=\bar{\eta}$ and then a feature vector is assumed to be drawn according to a density $f_y(x)$. Unlabeled data is thus drawn from the mixture $\eta f_1+\bar{\eta} f_2$. The second key assumption is that the class of mixture models is identifiable, i.e., we can infer the mixture model uniquely given only unlabeled data.
After observing enough unlabeled data to identify the mixture, we merely have to figure out how to label each part of the two mixture components. As we thus have only to decide between two alternatives, we can find this classifier $h$ by a simple likelihood ratio test, which converges exponentially fast to the Bayes risk in the number of the labeled samples $n$:
\begin{equation}
R(h)-\min\limits_{h\in H} R(h) 
\leq  \exp \left(n \ln (2 \sqrt{\mu \bar{\mu}} \int\sqrt{f_1(x)f_2(x)dx} )+o(n) \right)
\end{equation}

For the analysis it is necessary to assume that one has an infinite amount of unlabeled data.
The work is continued in \citep{Castelli2}, where the authors consider cases where we already have knowledge about the densities $f_y$. \citet{Sinha} extend a similar framework to the case where the marginal distribution $P(x)$ is unknown. They assume instead that $P(x)$ can be well estimated with a mixture of two spherical gaussian distributions with density functions $f_1(x)$ and $f_{-1}(x)$. In particular they assume that
\[
||f_1-P( \cdot | Y=1)||_S
\]
and
\[
|| f_{-1}-P( \cdot | y=-1)||_S,
\]
where $|| \cdot ||_S$ is a Sobolev norm, can be bounded with a small number.

In a way this work actually ties in with the impossibility result from Section \ref{seeger}. There we presented work that showed, that under a specific data generating process semi-supervised learning is in a way impossible. In essence one assumed that one draws first the feature $x$, and then dependent on $x$ a label $y$, see also Figure \ref{Datagen}. The work we presented now actually flipped the data generating process, i.e. we first draw $y$ and dependent on that we draw $x$. In conclusion one can see the analysis presented above as an example of a successful semi-supervised learner, for a data generating process with arrows reversed from Figure \ref{Datagen}.

\section{Assuming Classes are Clustered and Separated} \label{cluster1}

In \citep{Rigollet}, we are presented with explicit bounds on the generalization error using an alternative formulation of the cluster assumption. The approach closely resembles the work described in the previous section and, similarly, enables us to come to exponentially fast convergence under semi-supervision.

The work's initial, elementary setup is that we are given a collection of pairwise disjoint clusters $C_1, C_2,\ldots$ based on which we make a \emph{cluster assumption}, i.e, we assume that the optimal labeling function
\[
x \mapsto \sign(P(Y=1|X=x)-\frac{1}{2})
\]
is constant on each cluster $C_i$. So the clusters have a label-purity of some degree, which we can express as follows:
\begin{equation} \label{purity}
\delta_i=\int_{C_i} |2P(Y=1|X=x)-1| dP(x).
\end{equation}
The cluster $C_i$ is called pure if and only if $\delta_i$ is $1$. 

Assuming now that we know the clusters, we let $h^{\text{\normalfont SSL}}_n(x)$ be the majority voting classifier per cluster. More formally, given a labeled sample $S_n$ let
\[
X_i^{+}:=\{(x,y) \in S_n \ | \ x \in C_i, y=1\}
\]
and similarly
\[
X_i^{-}:=\{(x,y) \in S_n \ | \ x \in C_i, y=-1\}.
\]
Then given a new data point $x \in C_i$ we set
\begin{equation}
h^{\text{\normalfont SSL}}(x)=
\begin{cases}
1 & \text{if \ }  |X_i^{+}| \geq |X_i^{-}| \\
-1 & \text{if \ }  |X_i^{+}| < |X_i^{-}|.
\end{cases}
\end{equation}
Note that this defines only a function on the clusters. The paper argues, however, that unlabeled data cannot help where no unlabeled data was observed. Consequently it only analyses the possible gain from unlabeled data on the clusters. Thus the excess risk considered can now be restricted to the set $C:=\bigcup C_i$ and so we consider the risk
$$\mathcal{E}_C(h)=\int_C |2P(Y=1|X=x)-1| I_{\{h(x) \neq h^*(x) \}}dP(x), $$
where $h^*$ is the Bayes classifier.  The following theorem then expresses the gain one can make with respect to the expected cluster excess risk.
\begin{theorem} [{\citet[Theorem 3.1]{Rigollet}}]
Let $(C_i)_{i \in I}$ be a collection of sets with $C_i \subset \mathcal{X}$ for all $i \in I$ such that this collection fulfills the above defined cluster assumption. Then the majority voting classifier $h^{\text{\normalfont SSL}}_n$ as defined above satisfies
\begin{equation} \label{exponential}
\mathbb{E}_{S_n,U_m}\left[ \mathcal{E}_C(h^{\text{\normalfont SSL}}_n) \right] \leq 2 \sum\limits_{i \in I} \delta_i e^{\frac{-n \delta^2_i}{2}}.
\end{equation}
\end{theorem}
In other words, knowing the clusters, we recover the exponential convergence in the labeled sample size as in \citep{Castelli}. There is, however, a very important remark to be made here. This analysis goes only through if $\delta_i \neq 0$, which is the case if almost all $x$ in cluster $i$ have $P(Y=1 \mid X=x)=\frac{1}{2}$. So implicitly we assume that the posterior probability distribution is bounded away from $\frac{1}{2}$. In other words, the analysis does not hold uniformly over all distributions, as usually wanted for sample complexity bounds. But this distributional assumption is part of the \emph{strong} assumption which is needed to derive the result.

The biggest effort of the paper goes into the definition of clusters and the finite sample size estimation of such. The derivations are rather extensive and, as in many other parts of the review, we limit ourselves here to a description of the underlying intuition. To start with, one assumes that the marginal distribution $P(X)$ allows for a density function $p(x)$ with respect to the Lebesgue measure. With that one can define the density level sets of $\mathcal{X}$ w.r.t. a parameter $\lambda >0$ as
\[
\Gamma(\lambda):=\{ x \in \mathcal{X} \ | \ p(x) \geq \lambda \}.
\]
For a fixed $\lambda >0$, we think of a clustering essentially as path-connected components of the density level sets $\Gamma(\lambda)$, where it is ensured that pathological cases are excluded. Estimating the set $\Gamma(\lambda)$ with finitely many unlabeled samples adds a slack term to Inequality \eqref{exponential} that drops polynomially in the unlabeled sample size. Therefore, to ensure that we stiast l can learn exponentially fast, the number of unlabeled samples has to grow exponentially with the number of labeled samples.

\section{Assuming Classes are Clustered but not Necessarily Separated} \label{cluster2}

\begin{figure}[t!]
\centering
  \subfloat[The clusters $C_1$ and $C_2$ are separated with margin $\gamma$. The different decision regions are here just the clusters.]{%
    \includegraphics[width=0.3\textwidth,valign=t]{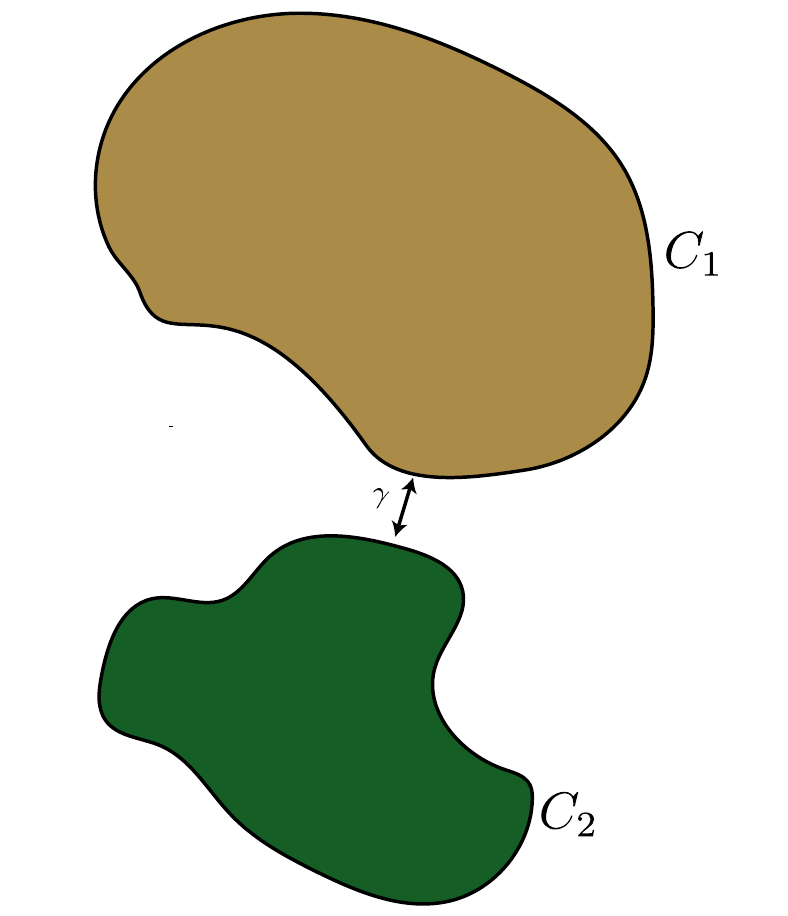}%
    \vphantom{\includegraphics[width=0.3\textwidth,valign=t]{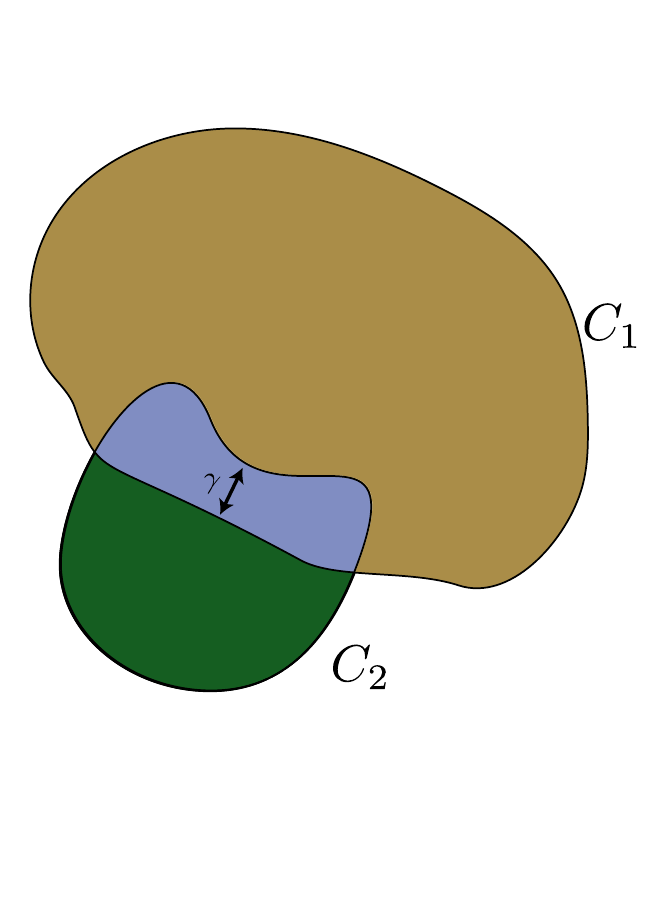}}

  } \quad
      \hphantom{\includegraphics[width=0.1\textwidth,valign=t]{Margin_Negative_small}}
  \subfloat[The clusters $C_1$ and $C_2$ are have an overlap (light blue) with margin $\gamma$. The three colors also constitute three different decision sets.]
  {%
    \includegraphics[width=0.3\textwidth,valign=t]{Margin_Negative_small}%
  }
  \caption{Picture (a) shows the concept of a positive $\gamma$-margin, while (b) shows a negative $\gamma$-margin.}
  \label{margins}
\end{figure}

\citet{Singh} propose yet another formalization of the cluster assumption.  More specifically, it is one that allows to distinguish cases where SSL does help and where not. This is achieved by restricting the class of distributions $\mathcal{P}$ and then investigating which of those distributions allow for successful semi-supervised learning. The class $\mathcal{P}$ is constructed such that the marginal distributions constitute of different clusters that are at times easy to distinguish and in other cases not. The marginal densities  $p(x)$ from $\mathcal{P}$  are given by mixtures of $K$ densities $p_k$. That is,
\[
p(x)=\sum_{i=1}^{K}a_k p_k(x)
\]
with $a_k >0$ and $\sum_{i=1}^{K}a_k=1$ and each $p_k$ has support on a set $C_k \subset \mathcal{X}$ which fulfills particular regularity conditions. We refer to these sets $C_k$ as clusters and each of these is assumed to have its own smooth label distribution function $p_k(y|x)$. So with probability $a_k$ we draw from $p_k(x)$ and then label $x$ according to $p_k(y|x)$. We further only consider distributions that lead to clusters with margin, with our without overlap, of at least $\gamma$ (see also Figure \ref{margins}), and denote the resulting class of distributions by $\mathcal{P}(\gamma)$. 
In this formulation, the clusters are not the main interest, but rather what the authors call the \emph{decision sets}.

To define a decision set, we take $C_k^c$ to be the complement of $C_k$ and, in addition, define $C_k^{\neg c}:=C_k$. Now, a set $D \subset \mathcal{X}$ is called a decision set if it can be written as $$ D=\bigcap\limits_{k \in K} C_k^{i_k}$$ with $i_k \in \{c, \neg c\}$ for all $k \in K$.  See Figure \ref{margins} (b) for an example. The advantage of decision sets over clusters is that the full distribution $p(x,y)$ is not necessarily smooth on each cluster, as they might exhibit jumps at the borders. On the decision sets, however, $p(x,y)$ is smooth as long as each $p_k(y|x)$ is smooth. Thus, if we would know the decision sets we could use a semi-supervised learner that uses the smoothness assumption.

The main theorem answers the question whether or not one can learn the decision sets from finitely many unlabeled points.
\begin{theorem} [{\citet[Corollary 1]{Singh}}]
Let
\[
\mathcal{E}(h)=R(h)-R^*
\]
be the excess risk with respect to the Bayes classifier $R^*$. Assume that $\mathcal{E}$ is bounded and that there is a learner $h^D_n$ that has knowledge of all decision sets $D$ and, additionally, fulfills the following excess risk bound:
\begin{equation}
\sup\limits_{P \in \mathcal{P}(\gamma)} \mathbb{E}_P[ \mathcal{E}(h^D_n)] \leq \epsilon_2(n).
\end{equation}
Assume that
\[
|\gamma| > 6 \sqrt{d} \kappa_0 \left( \frac{(\ln m)^2}{m}\right)^{\frac{1}{d}},
\]
where $\kappa_0$ is a constant, then there exists a semi-supervised learner $h^{\text{\normalfont SSL}}_{n,m}$ such that
\begin{equation} \label{decisionsetsssl}
\sup\limits_{P \in \mathcal{P}(\gamma)} \mathbb{E}_P[ \mathcal{E}(h^{SS}_{n,m})] \leq \epsilon_2(n)+O\left(\frac{1}{m}+n \left( \frac{(\ln m)^2}{m}\right)^{\frac{1}{d}}\right).
\end{equation}
\end{theorem}

We immediately note the following. If the learner $h^D_n$ that knows the decision sets has a convergence rate of $\epsilon_2(n)$, it follows from Inequality \eqref{decisionsetsssl} that the unlabeled data needs to increase with a rate of $\epsilon_2(\frac{1}{n})$ to ensure that the semi-supervised learner has the same convergence rate as $h^D_n$. For example, if $h^D_n$ converges exponentially fast, we need exponentially more unlabeled than labeled data, which exactly corresponds to the finding in the previous section.

All in all, the intuition behind the theorem is fairly clear. The bigger $\gamma$, the less unlabeled samples we need to estimate the decision sets $D$.  Moreover, once we know those sets, we can perform as well as $h^D_n$.  Now, to analyze if a semi-supervised learner that first learns the decision sets empirically has an advantage over all supervised learners, we first find minimax lower bounds for all fully supervised learners. We can then give upper bounds for a specific semi-supervised learner and the conclusions follow easily: for SSL to be useful, the parameter $\gamma$ and the number of unlabeled samples should be such that the fully supervised learner cannot distinguish the decision sets, while the semi-supervised learner can. As a consequence, $\gamma$ should not be too big, because then the supervised learner can also distinguish the decision sets. Additionally, the unlabeled data should of course not be too little, for then the semi-supervised learner cannot distinguish the decision sets either.

To present specific differences between SSL and SL the authors assume that $\mathcal{X}=[0,1]^d$ and that the conditional expectations
\[
\mathbb{E}_{Y \sim p_k(Y|X=x)}[Y|X=x]
\]
are H{\"o}lder-$\alpha$ smooth functions in $x$. Depending on $\gamma$, the paper presents a table for cases when SSL can be essentially faster than SL. In those cases, the SL has an expected lower bound for the convergence rate of ${n^{-\frac{1}{d}}}$ while the convergence rate of the semi-supervised learner is upper bounded by ${n^{-\frac{2 \alpha}{2\alpha+d}}}$.

\section{Assuming Regression is Smooth Along a Manifold}\label{densitysensitive}

As we elaborate further in the discussion chapter, an issue in SSL is that most methods are based on assumptions on the full distribution. The core problem is that we usually cannot verify whether such assumptions hold or not. This is crucial to know, since in case the assumption does not hold, it is quite likely that we want to use a supervised learner instead. The work of \citep{Azizyan} is one of the few papers that touches on this topic and introduces a semi-supervised learner that depends on a parameter $\alpha$, where in case of $\alpha=0$ a purely supervised learner is recovered. The paper then gives generalization bounds for the semi-supervised learner when we cross-validate $\alpha$.  As this work uses the regression setting, while most other presented papers deal with classification, and gives a formalization of the manifold assumption, we present here the details. 

The authors use a version of the manifold assumption, so we enforce our estimated regression function $h^{\text{\normalfont SSL}}(x)$ to behave smoothly in high density regions. The density of the marginal distribution $P(X)$ is measured with a smoothed density function $p_\sigma(x)$
\begin{equation} \label{newdensity}
p_{\sigma}(x):= \int \frac{1}{\sigma^d} K\left(\frac{||x-u||}{\sigma}\right)dP(u),
\end{equation}
where $K$ is a symmetric kernel on $\mathbb{R}^d$ with compact support and $\sigma>0$. Let $\Gamma(x_1,x_2)$ be the set of all continuous paths
\[
\gamma:[0,L(\gamma)] \to \mathbb{R}^d
\]
from $x_1 \in \mathbb{R}$ to $x_2 \in \mathbb{R}$ with unit speed and where $L(\gamma)$ is the length of $\gamma$. With this we can define a new metric on $\mathbb{R}^d$, i.e., the so-called exponential metric, that depends on a parameter $\alpha \geq 0$ and the smoothed density $p_{\sigma}(x)$:
\begin{equation} \label{expmetric}
D(x_1,x_2)=\inf\limits_{\gamma \in \Gamma} \int_0^{L(\gamma)} e^{-\alpha p_{\sigma}(\gamma(t))} dt
\end{equation}
First, note that $\alpha=0$ corresponds to the Euclidean distance. Second, note that high values of $p_{\sigma}(x)$ on the path between two points $x_1$ and $x_2$ lead to shorter distances between those points in the new metric. This behavior gets of course more emphasized with large $\alpha$. If we assume that $Q$ is another kernel and we set $Q_{\tau}(x):=\frac{1}{\tau^d}Q(\frac{x}{\tau})$ we can define a semi-supervised estimator as follows:
\begin{equation} \label{densitysensiSSL}
h^{\text{\normalfont SSL}}(x):=\frac{\sum_{i=1}^n y_i Q_{\tau}(\hat{D}(x,x_i))}{\sum_{i=1}^n  Q_{\tau}(\hat{D}(x,x_i))}.
\end{equation}
The estimator is thus a nearest-neighbor regressor, where neighbors are weighted according to their distance in term of the previously defined $D$-metric.

The following theorems gives bounds on the squared risk of $h^{\text{\normalfont SSL}}$ under the assumption that 
\[
\sup\limits_{y \in \mathcal{Y}} |y| =M < \infty.
\]
In our formulation, the theorem depends on some unspecified regularity conditions and additional notation to which we return with more details shortly.
\begin{theorem} [{\citet[Theorem 4.1]{Azizyan}}] \label{densitysensithm}
Let $\mathcal{P}(\alpha,\sigma,L)$ be a class of probability measures that fulfill certain regularities depending on parameters $\alpha,\sigma,L \geq 0$. Assume that for all $P \in \mathcal{P}$, we have
\[
P(||\hat{p}_{\sigma}-p_{\sigma}|| \geq \epsilon_m) \leq \frac{1}{m},
\]
then
\begin{equation}\label{densitysensirisk}
\mathbb{E}_{S_n,U_m}[R(h^{\text{\normalfont SSL}}]  \leq  L^2(\tau e^{\alpha \epsilon_m})^2+\frac{1}{n}M^2(2+\frac{1}{e}) \mathcal{N}_{P,\alpha,\sigma}( e^{-\alpha \epsilon_m}\frac{\tau}{2})+\frac{4M^2}{m}. 
\end{equation}
\end{theorem}
In this formulation of the result, $\mathcal{N}_{P,\alpha,\sigma}(\epsilon)$ is the \emph{covering number} of $P$ in the $D$-metric, i.e., the minimum number of closed balls in $\mathcal{X}$ of size $\epsilon$ (w.r.t to the $D$-metric) necessary to cover the support of $P(X)$ \citep[see also][Chapter 27]{Shalev}. In the Euclidean case, when $\alpha=0$, we can bound
\[
\mathcal{N}_{P,\alpha,\sigma}(\epsilon) \leq (\frac{C}{\epsilon})^d
\]
with the help of a constant $C$. The covering number can be much smaller when $\alpha>0$ and $P(X)$ is concentrated on a manifold with dimension smaller than $d$.

The previous theorem may be difficult to grasp in full at a first read and the paper offers, under some further regularity conditions, a simplified corollary in addition.
\begin{corollary} [{\citet[Corollary 4.2]{Azizyan}}]
Assume that
\[
\mathcal{N}_{P,\alpha,\sigma}(\delta) \leq (\frac{C}{\delta})^{\xi}
\]
for a certain range of $\delta$. Furthermore, assume that $m$ is large enough and that $\tau(n,\alpha,\epsilon_m,\xi)$ is well chosen. Then for all $P \in \mathcal{P}(\alpha,\sigma,L)$
\begin{equation}
\mathbb{E}_{S_n,U_m} [R(h^{\text{\normalfont SSL}}] \leq \left(\frac{C}{n} \right)^{\frac{2}{2+\xi}}.
\end{equation}
\end{corollary}

Following this, the paper analyzes the additional penalty one occurs by trying to find the best $\alpha$. We start by discretizing the parameter space $\Theta=\mathcal{T} \times \mathcal{A} \times \Sigma$ such that $\theta=(\tau, \alpha, \sigma) \in \Theta$ and $| \Theta |=J < \infty$. Assume now that we have, in addition to the training sample $S_n$, also a validation set
\[
V=\{(v_1,z_1),\ldots,(v_n,z_n) \},
\]
which, for convenience, is also of size $n$. Let $h^{\text{\normalfont SSL}}_{\theta}$ be the semi-supervised hypothesis trained on $S_n$ with the parameters $\theta$. We then choose the final hypothesis $h^{\text{SSL}}$ by optimizing for $\theta$ through cross-validation
\begin{equation} \label{estimatorcrossval}
h^{\text{\normalfont SSL}}:=\arg \min \limits_{h^{\text{\normalfont SSL}}_{\theta}} \sum_{i=1}^{n} (h^{\text{\normalfont SSL}}_{\theta} (v_i)-z_i)^2.
\end{equation}

\begin{theorem} [{\citet[Theorem 6.1]{Azizyan}}]
Let
\[
\mathcal{E}(h):=R(h)-R(h^*)
\]
be the excess risk, where $h^*$ is the true regression function. There exist constants\footnote{It should be noted that these are not universal. They depend to some degree on the problem at hand.} $0<a<1$ and
\[
0<t<\frac{15}{38(M^2+\sigma^2)}
\]
such that
\begin{equation} \label{crossvalineq}
 \mathbb{E}_{S_n,U_m,V}[\mathcal{E}(h^{\text{\normalfont SSL}})] \\ \leq  \frac{1}{1-a}\left( \min\limits_{\theta \in \Theta} \mathbb{E}_{S_n,U_m}[\mathcal{E}(h^{\text{\normalfont SSL}}_{\theta})]+\frac{\ln(nt4M^2)+t(1-a)}{nt}\right).
\end{equation}
\end{theorem}

This result is particularly interesting since we can always compare implicitly to the supervised solution, as long as we include $\alpha=0 \in \mathcal{A}$. From Inequality \eqref{crossvalineq} we see that the validation process introduces a penalty term of size $O(\frac{\ln(n)}{n})$. This of course allows us to flexible choose between the semi-supervised and the supervised method.

In a final contribution, the authors identify a case where the semi-supervised learning rate can be strictly better than the supervised learning rate. The setting considered is much like the one we have seen in Section \ref{effectivessl}. In particular, they construct a set of distributions $\mathcal{P}_n$, which depends on the number of labeled samples, such that
\begin{enumerate}
\item the estimator $h^{\text{\normalfont SSL}}(x)_{\tau,\alpha,\sigma}$, as defined in Equation \eqref{densitysensiSSL}, fulfills $$\sup\limits_{P \in \mathcal{P}_n} \mathbb{E}_{S_n} [R(\hat{f}_{\tau,\alpha,\sigma})] \leq \left(\frac{C}{n} \right)^{\frac{2}{2+\xi}},$$
under the assumption that $m \geq 2^{\frac{2}{2+\xi}}$;
\item for all purely supervised estimators $h^{\text{SL}}$ we have that
$$\sup\limits_{P \in \mathcal{P}_n} \mathbb{E}_{S_n} [h^{\text{SL}}] \geq \left(\frac{C}{n} \right)^{\frac{2}{d-1}}.$$
\end{enumerate}
To obtain essentially different learning rates we need that $\xi < d-3$, which is the case if $P$ is concentrated on a set with dimension strictly less than $d-3$ \citep[Lemma 1]{Azizyan}. It is also worth noting that the construction of $\mathcal{P}_n$ works by concentrating the distributions more for bigger $n$. If $\mathcal{P}_n$ does not concentrate, and remains smooth for bigger $n$, the labeled data is already enough to approximate the marginal distribution.

This is similar to the work presented in Section \ref{cluster2}, as they also show that SSL can only work if the marginal distribution $P(X)$ is not too easy to identify. We can also draw parallels to the work presented in Section \ref{niyogi}; if we would restrict the domain distributions such that only smooth circle embeddings would be allowed, a supervised learner could also learn efficiently. This is because then a finite number of labeled samples would be sufficient to learn the domain distribution uniformly, so the semi-supervised learner would lose its benefits.

\chapter{Learning in the Transductive Setting}\label{Transductive}

While most SSL methods use unlabeled data to try and find better inductive classification rules, i.e., rules that apply to the whole input domain $\mathcal{X}$. Some works consider schemes where one only cares about the labels of the unlabeled data specifically at hand. Such methods are often referred to as transductive and they have been argued to be an essential step forward compared to inductive methods, in particular by Vapnik (see for instance  \cite[Chapter 8]{Vapnik1998} and \cite[Chapter 1.2.4 and 25]{Chapelle}). We review the most important theoretical results in the context of this survey. A more detailed overview of theoretical and practical transductive learning can be found in Chapter 2 of \citep{Pechyony}. In Subsection \ref{transbounds}, we present learning bounds that apply specifically to this transductive setting, though they often arise as direct extensions to the \emph{supervised} inductive case and related work. In Section \ref{safelearning}, we present two papers that touch on the topic of so-called safe semi-supervised learners\footnote{Incidentally, this is a topic that is not covered by \citet{Pechyony}}. The aim of these methods is to construct semi-supervised learners that are never worse than their supervised counterparts.

One essential difference, based on which two distinct transductive settings can be identified, is the way the sampling of the labeled and unlabeled data comes about.
\begin{itemize}
\item[Setting 1]
\begin{enumerate}
\item We start with a fixed set of points $X_{n+m}=\{x_1,\ldots,x_{n+m}\}$.
\item We reveal the labels $Y_n$ of a subset $X_{n} \subset X_{n+m}$, which is uniformly selected at random. For notational convenience and without loss of generality, we usually assume that $X_n$ are the first $n$ and $X_m$ are the last $m$ points of $X_{n+m}$.
\item Based on $S_n=(X_n,Y_n)$ and $X_m$  we aim to find a classifier $h$ with good performance as given by 
\[
R_m(h):=\sum_{i=n+1}^{n+m} l(x_i,y_i).
\]
\end{enumerate}
\item[Setting 2]
\begin{enumerate}
\item We start with a fixed distribution $P$ on $\mathcal{X} \times \mathcal{Y}$.
\item We draw $n$ i.i.d. samples according to $P$ to obtain a training set $S_n$. We draw an additional $m$ i.i.d. samples according to $P(X)$ to obtain a test set $X_m$.
\item Based on $S_n=(X_n,Y_n)$ and $X_m$ we try to find a classifier $h$ with good performance as specified by 
\[
\mathbb{E}_{S_n,X_m} \left[ \frac{1}{m} \sum_{i=n+1}^{n+m} l(h(x_i),y_i) \right].
\]
\end{enumerate}
\end{itemize}
Note that in the first setting, we basically sample without replacement and, as an result, the samples are dependent. The work concerning transductive learning that we present here deals with Setting 1.  This is primarily out of convenience, but we note that one can always transform bounds from Setting 1 to bounds in Setting 2 \citep[Theorem 8.1]{Vapnik1998}.

We remark that in this section our test error is denoted by $R_m(h)$ and the training error by $R_n(h)$. This reflects that the test is of size $m$ while the training set of size $n$. We do not use the hat notation here, as in the transductive setting, we do not necessarily have an underlying distribution.

\section{Transductive Learning Bounds}\label{transbounds}
In this section we present some transductive learning bounds, which often arise as an extension of their indudctive counter parts. Generally we find that the rates are in some order of $\frac{1}{\min(m,n)}$. At a first glance, that may seem somewhat surprising considering that the labeled and unlabeled data play an equivalent role in terms of convergence. While slow convergence for $n\ll m$ is not really surprising, given that we have few labeled data points to train on, one has to realize that in the case where $m\ll n$ the transductive risk has a very high variance and thus we have large intervals for high-confidence estimations. In this sense we see that estimating a transductive risk is essentially different than estimating the risk over a distribution.

\subsection{Vapnik's Implicit Transductive Bound}

The study of transductive inference goes back at least to the original work by Vapnik \cite[see, for example,][]{Vapnik1982}. In this section, our primary source is \cite{Vapnik1998} and we mainly consider the result found as Equation (8.15) in Theorem 8.2 of that work.

Assume that we are given $n+m$ samples and we take at random $n$ samples on which we can train. We then want to estimate the error we make on the leftover $m$ samples. Vapnik shows that a hypergeometric distribution describes the probability that the observed error on the train and test set is bigger than $\epsilon$  $$P\left( \frac{|{R}_m(h)-{R}_n(h)|}{\sqrt{{R}_{n+m}(h)}}>\epsilon\right).$$ Let $\epsilon^*$ be the smallest $\epsilon>0$ such that $$P\left( \frac{|{R}_m(h)-{R}_n(h)|}{\sqrt{{R}_{n+m}(h)}}>\epsilon \}\right) \leq 1-\delta.$$ Using a uniform bound\footnote{Note that in the transductive case we effectively can have only finitely many different hypotheses.} and substituting
\[
{R}_{n+m}=\frac{m}{n+m}{R}_m+\frac{n}{n+m}{R}_n
\]
one can derive the following result.
\begin{theorem}
For all $h \in \{-1,1\}^{n+m}$, the following inequality holds with a probability of $1-\delta$:
\begin{equation}  \label{vapnikimplicit}
 {R}_{m}(h) \leq {R}(h)+\frac{(\epsilon^*)^2m}{2(m+n)}+\epsilon^*\sqrt{{R}(h)+\left(\frac{\epsilon^*m}{2(m+n)}\right)^2}
\end{equation}
\end{theorem}

A core problem with this inequality is that the term $\epsilon^*$ is an implicit function of $n,m,\delta$ and $h$ and thus it is unclear what the learning rates are that we can actually achieve. This issue is addressed in the paper covered in the next subsection.

\subsection{Bounds as a Direct Extension of Inductive Bounds }

The transductive bound of Inequality \eqref{vapnikimplicit} is difficult to interpret as it contains a function which can only be implicitly calculated. \citet{Derbeko} find explicit transductive bounds in a PAC-Bayes framework. We present a bound from the paper which is essentially a direct extension of a supervised inductive bound by \citet{McAllester1}. To come to their result, they consider a Gibbs classifier, which we first introduce.

Let $q$ be any distribution over the hypothesis set $H$. The Gibbs classifier $G_q$ classifies a new instance $x \in \mathcal{X}$ with an $h \in H$ drawn according to $q$. The risk of $G_q$ over the set $S_n$ is then
\[
{R}_n(G_q)=\mathbb{E}_{h \sim q}[ \frac{1}{n}\sum_{i=1}^n l(h(x_i),y_i)].
\]
\begin{theorem} [{\citet[Theorem 17]{Derbeko}}]
Let $p$ be any (prior) distribution on $H$, which may depend on $S_{n+m}$, and let $\delta>0$. Then for any randomly selected subset $S_n \subset S_{n+m}$ and for any distribution $q$ on $H$, it holds with probability at least $1-\delta$ that
\begin{equation}  \label{Pacgibbs}
 {R}_m(G_p) \leq {R}_n(G_p) + 
+  \frac{m+n}{m}\left(\sqrt{\frac{2{R}_n(G_p)(\kl(q||p)+\ln\frac{n}{\delta})}{n-1}} +\frac{2(\kl(q||p)+\ln\frac{n}{\delta})}{n-1}\right).
\end{equation}
\end{theorem}

This theorem is indeed a direct extension of the inductive supervised case as found under Equation (6) in \citep{McAllester1}. The only difference is that the term $\frac{m+n}{m}$ is missing. Although \citet{McAllester2} show that under certain conditions one can select the prior $p$ after having seen $S_m$, this is generally not allowed in inductive PAC-Bayesian theory. In the transductive setting this is allowed, however, as we only care about the performance on the points from the set $S_{n+m}$. In a way this is the same as learning with a fixed distribution when our fixed distribution has only mass on finitely many points \citep{Benedek2}.

\citet{Derbeko} exploit the previous observation by choosing a prior $p$ with a cluster method. More precisely, after observing the dataset $(X_{n+m})$ one constructs $c$ different clusterings on it. Each clustering leads to multiple classifiers by assigning all points in a cluster to the same class. One then puts essentially a uniform prior $p$ on those classifiers and we select a posterior distribution $q$ over the classifiers by minimizing Inequality \eqref{Pacgibbs}, and obtain the Gibbs classifier $G_q$.

Comparing this approach to the fully supervised (and thus necessarily inductive) case, we realize that the possible performance improvements have the same flavor as the improvements one can gain in semi-supervised learning with assumptions, as analyzed in Chapters \ref{weakassumptions} and \ref{strongassumptions}. Using the clustering approach sketched above reduces the penalty in Inequality \eqref{Pacgibbs}, which is coming from $\kl(q||p)$. In other words: we reduce the variance of the classifier. Clearly, on the other hand, using a clustering approach biases our solution and we get degraded performance compared to a supervised solution if the clusterings have a high impurity, i.e., clusters do not have clear majority classes.

\subsection{Bounds Based on Stability}\label{stabilitysec}

In \cite{El-Yaniv}, transductive bounds are explored under the assumption of stability, i.e., the notion that the output of a classifier does not change much if we perturb the input a bit. The transductive bounds presented are an extension of the inductive bounds that use the notion of \emph{uniform stability} \citep[see][]{Bousquet} and \emph{weak stability} \citep[see][]{Kutin,Kutin2}). We cover the simpler transductive bound based on uniform stability and explain the difference to weak stability.

Assume that $h^{\text{trans}} \in H$ is a transductive learner.  That is, a hypothesis that we (deterministically) choose based on a labeled set $S_n$ and an unlabeled set $X_m$. Furthermore, define
\[
S^{ij}_n:=\left (S_n \setminus \{(x_i,y_i)\} \right) \cup \{(x_j,y_j)\}.
\]
So $S^{ij}_n$ is the set we obtain when we replace in $S_n$ the $i$-th example from the training set with the $j$-th example from the test set.  Similarly, define
\[
X_m^{ij}:=\left( X_m \setminus \{x_j\} \right) \cup \{x_i\}.
\]
We say that $h^{\text{trans}}$ is \emph{$\beta$-uniformly stable} if for all choices $S_n \subset S_{n+m}$ and for all $1\leq i,j \leq n+m$ such that $(x_i,y_i) \in S_n$ and $x_j \in X_m$ it holds that
\begin{equation} \label{uniformstability}
 \max\limits_{1 \leq k \leq n+m} |h^{\text{trans}}_{(S_n,X_m)}(x_k)-h^{\text{trans}}_{(S_n^{ij},X_m^{ij})}(x_k) | \leq \beta.
\end{equation}
In words: the transductive learner $h^{\text{trans}}$ is $\beta$-uniformly stable if the output changes less than $\beta$ if we exchange two points from the train and test set.

The bounds are formulated using a $\gamma$-margin loss. With $\gamma >0$, we define
\begin{equation} \label{gammamargin}
 l_{\gamma}(y_1,y_2)=\max(0, \min (1,1-\frac{y_1 y_2}{\gamma})).
\end{equation} Consequently, we can write $R_{\gamma}(h)$ for the risk of $h$ when measured with the loss $l_{\gamma}$. Note that for $\gamma \rightarrow 0$ the $l_{\gamma}$ loss converges to the 0-1 loss.
\begin{theorem} [{\citet[Theorem 1]{El-Yaniv}}]
Let $h^{\text{\normalfont trans}}$ be a $\beta$-uniformly stable transductive learner and $\gamma,\delta >0$. Then, with probability of at least $1-\delta$ over all train and test partitions, we have that
\begin{equation} \label{stabilitybound}
R_m(h^{\text{\normalfont trans}}) \leq  R^{\gamma}_n(h^{\text{\normalfont trans}}) 
+  \frac{1}{\gamma}O\left(\beta \sqrt{\frac{mn\ln\frac{1}{\delta}}{m+n}} \right)+O\left( \sqrt{(\frac{1}{m}+\frac{1}{n}) \ln \frac{1}{\delta}} \right).
\end{equation}
\end{theorem}

Note that $\beta$ is depended on $n$ and $m$ and we expect that the bigger our training set is, the less our algorithm changes if we exchange two samples from the train and test set. The transductive bounds based on Rademacher complexities, reviewed in the next section, can achieve convergence rates of $\frac{1}{\sqrt{\min(m,n)}}$. To obtain the same rate with Inequality \eqref{stabilitybound}, we need that $\beta$ behaves as $O\left(\sqrt{(\frac{1}{n}+\frac{1}{m})\frac{1}{\min(n,m)}} \right)$. This stability rate can indeed be achieved for regularized RKHS methods as demonstrated by \citet{Johnson3}.

\subsection{Bounds Based on Transductive Rademacher Complexities}\label{tRade}

Rademacher complexities are a well studied and established tool for risk bounds in the inductive case \citep{bartlett3}. \citet{El-Yaniv2} introduce a transductive version of these quantities. While in the inductive case, we have to chose our hypothesis class before seeing any data, the transductive case allows us to chose the hypothesis class data-dependent. The definition of the transductive Rademacher complexity of a hypothesis class $H$ closely follows the inductive case and is denoted by $\tRad(H)$. Utilizing the $\gamma$-margin loss function \eqref{gammamargin} and the corresponding empirical risk ${R}^{\gamma}(h)$, the paper shows then that for all $h \in H$, we have that
$$
R_{m}(h) \leq {R}^{\gamma}_n(h)+\frac{\tRad(H)}{\gamma}+O\left(\frac{1}{\sqrt{\min(m,n)}}\right).
$$
This bound can be used to directly estimate the transductive risk for transductive algorithms.

\citet{Maximov} make different use of Rademacher complexities in their derivation of risk bounds for a specific multi-class algorithm. Their algorithm uses a given clustering based on the full data to find a hypothesis which is in a certain way compatible with the clusters obtained. The transductive multi-class Rademacher complexities then make direct use of this clustering. With this algorithm the authors show that if we have $K$ initial classes one can achieve a learning rate in the order of $\tilde{O}(\frac{\sqrt{K}}{\sqrt{n}}+\frac{K^{3/2}}{\sqrt{m}})$ \citep[see][Corollary 4]{Maximov}.
Not surprisingly the learning rates are essentially the same as in the binary transductive cases.  We note, however, that the analysis was done within Setting 2.

\subsection{Bounds Based on Learning a Kernel}\label{Kernelmatrix}

As a direct extension of the inductive case \citep[see, for example,][]{Bartlett4}, \citet{Lanckriet} propose to use the unlabeled data to learn a kernel that is suitable for transductive learning. The  idea is to use a kernel method that allows to choose from a certain class of kernels in order to optimize the objective function. The presented PAC-bound shows that good (transductive) performance is achieved with a good trade-off between the complexity of the kernel class and the empirical error.

Their exemplary kernel classes are designed as follows. Given an initial set of kernels $\{K_1,\ldots,K_k\}$
they define $$K_c:=\{K=\sum_{j=1}^k \mu_j K_j \ |  K \succcurlyeq 0, \mu_j \in \mathbb{R}, \trace(K) \leq c \}$$
and
$$K^{+}_c:=\{K=\sum_{j=1}^k \mu_j K_j \ |  K \succcurlyeq 0, \mu_j \in \mathbb{R}, \mu_j \geq 0, \trace(K) \leq c \}.$$ Every class of kernels $\mathcal{K}$ give rise to the hypothesis set
\[
\begin{split}
& H_{\mathcal{K}} = \\
& \{h(x_j):=\sum_{j=1}^{2n} \alpha_i K_{ij} \ | \ K \in \mathcal{K}, \alpha=(\alpha_1,\ldots,\alpha_{2n}) \in \mathbb{R}^{2n}, \alpha^t K \alpha \leq \frac{1}{\gamma^2} \},
\end{split}
\]
from which we will chose a hypothesis.

 We now come to the paper's claim.  The original formulation of the theorem is rather long and contains some additional definitions and clarification as part of it.  In order to make the presentation a bit easier to access, we formulate the core result as a theorem, which should convey its basic structure and idea. Only afterwards, we will provide the missing details of the theorem.
\begin{theorem}[{\citet[Theorem 24]{Lanckriet}}] \label{Kerneltrans}
For every $\gamma >0$, with probability at least $1-\delta$ over every training and test set of size $n$ (so $m=n$), uniformly chosen from $(X,Y)$, we have for every function $h \in H_\mathcal{K}$ that
$$
R_{m}(h) \leq \hat{R}^{\hinge}_n(h) +\frac{1}{\sqrt{n}}\left(4+\sqrt{2\log (\frac{1}{\delta}}) + \sqrt{\frac{\comp(\mathcal{K})}{n \gamma^2}}\right).
$$
where $\hat{R}^{\hinge}(h)$ is the empirical hinge loss of $h$ and $\comp(\mathcal{K})$ is a complexity measure of $\mathcal{K}$.
\end{theorem}

This last measure of complexity, $\mathcal{K}$, is defined as
$$
\comp(\mathcal{K})= \mathbb{E} \max\limits_{K \in \mathcal{K}} \sigma^t K \sigma
$$
with $\sigma$ being a vector of $2n$ Rademacher variables.
For the previously defined kernel classes $\mathcal{K}_c$ and $\mathcal{K}^{+}_c$, this complexity measure can, in turn, be bounded by
$$
\mathcal{K}_c=c \mathbb{E} \max\limits_{K \in \mathcal{K}} \sigma^t \frac{K}{\trace{K}} \sigma \leq cn,
$$
and
$$
\mathcal{K}^{+}_c \leq c \min \left(k, n \max\limits_{1\leq j \leq k} \frac{\lambda_j}{\trace(K_j)}\right).
$$
In this last expression, $\lambda_j$ is the largest eigenvalue of $K_j$.

Since $m$ is taken to equal $n$, we find that the above bound gives the same learning rate $O(\frac{1}{\sqrt{m+n}})$ as we also found in Subsections \ref{stabilitysec} and \ref{tRade}. We would, however, not expect that the rate of $O(\frac{1}{\sqrt{m+n}})$ also holds for different choices of $m$ and $n$. We would rather expect to get a term in the order of $O(\frac{1}{\sqrt{\min(m,n)}})$ as also explained in the introduction of Section \ref{transbounds}.   

On another note, we point out that the effect the unlabeled data has on this procedure depends on the initial kernel guesses $\{K_1,\ldots,K_k\}$, but there is no informed choice of those kernels.  

A possible informed choice of the initial kernels is explained in \cite{kernel2}. Taking $\psi_i$ to be the $i$-th eigenvector of the graph Laplacian $L$, we can set $K_i=\psi_i {\psi_i}^t$. As described in \citep{Chapelle}, we can then enforce classifiers found by this procedure to be smooth along the data manifold, if we enforce that $\mu_i$ is small when the eigenvalue of $\psi_i$ is large. Similar results are obtained by \citet{Johnson}, where the biggest difference are the kernels that are used. Instead of using an initial set of kernels, \citet{Johnson} use the spectral decomposition of a given kernel and shrinks it, where the shrinkage depends on the unlabeled data.

\section{Safe Transductive Learning}\label{safelearning}

In the semi-supervised learning community it is well known that using a semi-supervised procedure often comes with a risk of performance degradation \citep{risks}. This problem leads some authors to ask the question whether it is possible to perform semi-supervised learning in a safe way, which means to say that one can guarantee that the semi-supervised learner is not worse than its supervised counterpart.  So far we primarily compared SSL and SL risk bounds. But, even if the assumptions underlying the risk bounds are true, a smaller bound still does not guarantee improvements.

We look specifically at the approaches by \citet{Li} and \citet{Loog}. The results from both works are based on a minimax formulation and show that, in certain settings, one can indeed get to guarantee performance improvements by using SSL. The analysis is done in transductive Setting 1, which means that we have a training set $S_n$ and a test set $X_m$.

\subsection{A Minimax Approach for SVMs}

The baseline for the model proposed by \citet{Li} is the $S3VM$ \citep{Bennett}, which takes the unlabeled data into account by finding a large-margin solution. The proposed model $S4VM$ finds a few diverse proposal large-margin solutions, and then picks amongst these by means of a minimax framework to hedge against possible worst case scenarios. The idea is that, given that we found a set of a few potential solutions
\[
H_p=\{ h_1,\ldots,h_T \},
\]
we compare those solutions to $h^{SVM}$ and then choose the one with the biggest gain over $h^{SVM}$ within a minimax framework.

Assume for now that we know the true labels $Y_m=(y_{n},\ldots,y_{n+m})$ of $X_m$. With this we can calculate the gain and loss in performance when comparing the supervised $h^{SVM}$ to any other classifier $h$.
\begin{equation} \label{gain}
\text{gain}(h,Y_m,h^{SVM}):=\sum_{i=n}^{n+m} I_{\{h(x_i)=y_i\}}
I_{\{h^{SVM}(x_i) \neq y_i\}}
\end{equation}
\begin{equation} \label{loss}
\text{loss}(h,Y_m,h^{SVM}):=\sum_{i=n}^{n+m} I_{\{h(x_i) \neq y_i\}}
I_{\{h^{SVM}(x_i) = y_i\}}
\end{equation}
If we define our objective as to be the difference of those two, i.e.,
\begin{equation}
J(h,y,h^{SVM})=\text{gain}(h,Y_m,h^{SVM})-\text{loss}(h,Y_m,h^{SVM}),
\end{equation}
we can define a semi-supervised model $h^{\text{\normalfont SSL}}$ as the maximizer of this difference.  The problem is, of course, that we actually do not know the true labels. Therefore, let us assume a worst-case scenario, which leads us to the following max-min formulation:
\begin{equation}
h^{\text{\normalfont SSL}}=\arg \max\limits_{h \in H_p} \min\limits_{Y \in Y_p} J(h,Y,h^{SVM}).
\end{equation}
Here
\[
Y_p = \{ (h(u_1),\ldots,h(u_m)) \ | \ h \in H_p \}
\]
is the set of all possible labelings that we can achieve with $H_p$. To guarantee that our semi-supervised learner is not worse than the supervised learner it is important to assume that the true labels $Y_m$ are part of the set $Y_p$, because only then we can guarantee what follows.
\begin{theorem} [{\citet[Theorem 1]{Li}}] \label{safesvm}
If $Y_m \in Y_p$, the accuracy of $h^{\text{\normalfont SSL}}$ is never worse than the accuracy of $h^{SVM}$, when performance is measured on the unlabeled data $X_m$.
\end{theorem}

The crucial assumption is that $Y_m \in Y_p$, which corresponds in this case exactly to a low-density assumption. This is because the set $Y_p$ contains possible labelings that come from classifiers that fulfill the low density assumption. One can imagine to use the same procedure also for different assumptions as we can encode them through $Y_p$, i.e, the set of all labelings that we consider possible. While this paper still needs some assumptions, \citet{Loog} shows a case, covered in the next subsection, where we get guaranteed improvements assumption-free. This, however, comes at the cost of measuring the improvements in terms of likelihood, and not in terms of accuracy.

\subsection{A Minimax Approach for Generative Models} \label{mcpl}

The second technique taken from \citep{Loog} and also in the line of safe SSL research is, to our knowledge, the only approach to semi-supervised learning that considers a completely assumption-free setting. This comes at a cost of course, which we will expand on later.

The starting point is a family of probability density functions $p(x,y | \theta)$ on $\mathcal{X} \times \mathcal{Y}$, where $\theta \in \Theta$ is a parametrization. We then fix $\theta^{\text{SL}}$ to be the supervised maximum likelihood estimator for the model $p(x,y|\theta)$, i.e., $$\theta^{\text{SL}}=\arg \min\limits_{\theta \in \Theta} \left[\sum_{(x,y) \in S_n} \ln p(x,y | \theta)\right].$$
Let us assume for now that we know the true conditional probabilities
\[
p=(p_1,\ldots,p_{m+n} ) \in [0,1]^{m+n}
\]
with $p_i=p(1|x_i)$ for  $x_i \in S_n \cup X_m$. Indeed knowing this, we would rather optimize the expected log-likelihood of the model $p(x,y  |  \theta)$ evaluated on the complete dataset
\[
X_{n+m}=\{x_1,\ldots,x_{n+m}\}.
\]

This likelihood is given by
\begin{equation} \label{mcplcomplete}
L(\theta | X_{n+m},p)= \mathbb{E}_{Y \sim p} \left[\sum_{x \in X_{n+m}} \ln p(x,Y | \theta)\right].
\end{equation}
To be better than the supervised model $\theta^{\normalfont sup}$ on the complete (transductive) likelihood in Equtaion \eqref{mcplcomplete}, we would like to maximize the likelihood gain over it. In other words, we want to find the $\theta$ that maximizes the following difference in likelihood:
\begin{equation} \label{contrastive}
C(\theta,\theta^{\text{\normalfont SL}} | X_{n+m},p)=L(\theta | X_{n+m},p)-L(\theta^{\text{\normalfont SL}} | X_{n+m},p).
\end{equation}

We cannot maximize \eqref{contrastive} directly, as we do not know the class true probability distribution. Instead, we take $p(y_i | x_i)=1$ for all labeled points $(x_i,y_i) \in S_n$, which we denote by means of the vector
\[
p_n=(p(1|x_1),\ldots,p(1|x_n))
\]
and for the unlabeled points $X_m$, we consider worst case posteriors denoted by the $m$-vector $p_m$, which leads to the following max-min formulation:
\begin{equation} \label{MCPLest}
\theta^{\text{\normalfont SSL}}=\arg \max\limits_{\theta \in \Theta} \min\limits_{p_m \in [0,1]^{m}} C(\theta,\theta^{\text{\normalfont SL}} | X_{n+m},(p_n,p_m)).
\end{equation}
Note that the vector $p_m$ can be the true labels $Y_m$ of the unlabeled data $X_m$. Note also that
\[
C(\theta^{\text{\normalfont SSL}},\theta^{\text{\normalfont SL}}|X_{n+m},(p_n,p_m)) \geq 0
\]
for all $p_m \in [0,1]^{m}$, so in particular if $p_m=Y_m$, as we can always chose $\theta^{SSL}=\theta^{\text{\normalfont SL}}$. That means that the following theorem holds.
\begin{theorem} [{\citet[Lemma 1]{Loog}}]
Let $\theta^{\text{\normalfont SSL}}$ be a solution found in Equation \eqref{MCPLest}, then
\begin{equation} \label{MCPLresult}
L(\theta^{\text{\normalfont SL}} | X_{n+m},Y_{n+m}) \leq L(\theta^{\text{\normalfont SSL}}|X_{n+m},Y_{n+m}).
\end{equation}
\end{theorem}
Subsequently, \citet{Loog} shows that for some specific choices for the model $p(x,y  |  \theta)$, the previous inequality is strict almost surely, i.e., with probability 1. In that case, we are guaranteed that the transductive likelihood of our semi-supervised model is larger than that of the supervised model.

An important difference between this work and the one from previous subsection is that here one employs a generative model $p(x,y)$, while the SVM used by \citet{Li} is a discriminative model that inherently optimizes the class probability $p(y|x)$. The work by \citet{Jesse} (see also Subsection \ref{Jessepaper}) shows that, to some degree, it is actually necessary to use a generative model as the semi-supervised estimator of Equation \eqref{MCPLest} coincides with the supervised estimator for a large class of discriminative models. There are several explanations why a joint model $p(x,y)$ helps out in the situation. The intuitive and obvious one is that the likelihood of this model takes the marginal distribution $P(X)$ into account, which is a quantity that can be measured in part from unlabeled data.

\chapter{Discussion and Conclusion} \label{discussion}

We surveyed the main theoretical ideas and results that have been put forward over the past four decades in the field of semi-supervised learning.  Specifically, we focused on results that inform us about its potential and the possible lack of it. We covered the answers to the questions: What are the limits of semi-supervised learning? What are the assumptions of different methods? And what can we achieve if the assumptions are true?   We like to wrap up our survey and point out a few open problems and discussion points that become apparent in light of this survey.

\section{On the Limits of Assumption Free SSL}

In Section \ref{Impossibility}, we reviewed work that analyzes the limits of semi-supervised learning when no particular assumptions about the distribution are made that a semi-supervised learner can exploit. The most general formulation of this situation is captured in Conjectures \ref{bendavid1} and \ref{bendavid2}. They essentially state that a semi-supervised learner can beat all supervised learners by at most a constant. We then cover work that shows that the conjectures do actually not hold in full generality, but in particular situations only. They hold in particular for the realizable case in combination with hypothesis classes of finite VC-dimension, while they do not hold in the realizable or agnostic case for infinite VC-dimension. This leaves the case of agnostic PAC-learning with a finite VC-dimension as the situation that has not been fully settled yet and we identify this as a first important problem in SSL that is still open.

\section{How Good Can Constant Improvement Be?}\label{constantimprov}

The question studied in the previous subsection and Section \ref{constantimprove} is whether a semi-supervised learner can offer more than a constant improvement in terms of sample complexity. It seems a fair question to ask, however, how good a constant improvement can be.  Something that, certainly from a practical point of view, can still be very beneficial.  The answer to that can be seen through a thought experiment. Assume that we have two classes given by two concentric $d$-dimensional spheres.

Assume that we have enough unlabeled data for a manifold regularization scheme to identify the spheres. With this the semi-supervised learner needs only one labeled sample per class to give a perfect classification, while every supervised learner needs for good generalization a labeled sample size which increases in the dimension $d$. Although the manifold regularized classification needs only two samples, we also know that manifold regularization can only achieve constant improvement \citep{Mey}. This might seem contradictory, but this behavior is readily understood when we study the VC-dimensions. If the supervised classifier uses a hypothesis space $H$, we can interpret manifold regularization as switching to a restricted space $\tilde{H}_{\lambda}$. This space only contains hypotheses that fulfill a manifold assumption, where the regularization parameter $\lambda$ indicates to which degree this assumption is enforced. \citet{Mey} show that the improvement of using manifold regularization is at most $\vc(H) / \vc(\tilde{H}_{\lambda})$. If we set $\lambda$ high enough we can keep $\vc(\tilde{H}_{\lambda})$ constant, while $\vc(H)$ increases with the dimension $d$. This shows that the constant improvement can be arbitrarily high. While this example uses the manifold assumption, \citet{Golovnev} give an example with a semi-supervised learner that has the full knowledge of the domain distribution. We explain their particular example in Section \ref{constantimprove}.

All in all, this shows that the constant improvement can be arbitrarily high if we have further assumptions, e.g. the manifold assumption, or full knowledge of the marginal distribution. Still, a second important open problem that we can then identify is whether one can have arbitrarily high constants without assumptions and with limited unlabeled data.

\section{The Amount of Unlabeled Data We Need}

In Section \ref{effectivessl}, we treated three settings in which a semi-supervised learner can PAC-learn, while no supervised learner can. For that, we need, in principle, an infinite amount of unlabeled data. If a fixed finite amount of unlabeled data would be enough to learn under any given distribution $P$, we could just use the same strategy to learn in a supervised way, as we can always chose to ignore the label \citet[Theorem 1]{gopfert}. The way the examples of Section \ref{effectivessl} work is that for each fixed $P$ a finite, but arbitrarily large, amount of unlabeled data is sufficient. As a consequence, if we want to learn over all possible distributions, we need an arbitrarily large, i.e., infinite, amount of unlabeled data.

The improvements that semi-supervised learning can offer, which we present in Chapters \ref{Noassumptions}, \ref{weakassumptions} and \ref{strongassumptions}, do not necessarily need an infinite amount of unlabeled data, although this is sometimes assumed for convenience. The difference is that in those settings supervised learner are also able to PAC-learn, but a semi-supervised learner is able to do this with fewer labeled samples. In Sections \ref{cluster1} and \ref{cluster2} we saw two instantiations of a cluster assumption, and the authors showed that the amount of unlabeled data needs to increase exponentially with the amount of labeled data to make use of this assumption. This is because the error in finding the clusters decreases only polynomially in the number of unlabeled points as shown in Inequality \ref{decisionsetsssl}.

Working with a limited amount of unlabeled data remains a problematic topic, and given the previous results we believe that any limitation on unlabeled data will inhibit us from proving results that hold uniformly over all data distributions. Identifying settings where a limited amount of unlabeled data leads to large (constant) improvements remains a challenging problem.

\section{Using Assumptions in Semi-Supervised Learning} \label{assumptionsection}

In Chapters \ref{weakassumptions} and \ref{strongassumptions}, we investigated what a semi-supervised learner can achieve once assumptions are made, such as the ones we introduced in Section \ref{ssassumptions}. A semi-supervised assumption is a link between the domain distribution and the labeling function. In particular, we assume that we can ignore certain labeling functions after we have seen a specific domain distribution. The cluster assumption, for example, would exclude labeling functions that do not assign the same label to points belonging to the same cluster.  The obvious, but real problem with this is that we do not know if such assumptions do hold or not.  Clearly, one may be able to test the validity of certain assumption, but we would conjecture that testing for an assumption probably consumes as many labeled points as learning directly a good classification rule with a supervised learner.  In other words, the test would defy it purpose.  

To make this statement a bit more precise, let us define an assumption as a property of the distribution $P$ on $\mathcal{X} \times \mathcal{Y}$. Let $\mathcal{P}^A$ be a set of distributions on $\mathcal{X} \times \mathcal{Y}$. We say that $P$ fulfills assumption $A$ if and only if $P \in \mathcal{P}^A$. For example $\mathcal{P}^A$ could only contain distributions such that the marginal distributions $P(X)$ have always support on clusters and each cluster has a unique label. 
The important thing to note is, that the assumption $A$ is a property on $P$, so we need labeled samples to test whether its true or not. It is thus of interest to compare the consumption of labeled data for reducing the uncertainty about the assumption to the consumption of labeled data for the convergence of the semi-supervised learner. We might of course know a priori that the assumption is true and do not need to test it, but what if not?

One of the few works that analyze this is reviewed in Section \ref{densitysensitive}. \citet{Azizyan} show that one can get essentially faster rates if the assumption is true, but we pay a penalty of $O( \frac{\ln(n)}{n})$ if it is not true. \citet{propertytesting} investigates how one can test for a property in an active way, so when we can choose which samples we want to label. The implications of this testing procedure for semi-supervised learning are, at this point, a further open research question. Of course, one could insist that it is just not necessary to test whether an assumption is true or not. Following Vapnik's motto, we may want to avoid any intermediate form of testing to decide if an assumption is true or not, when, ultimately, we are merely interested in whether the semi-supervised learner performs better or not.  



\appendix

\chapter{Some Definitions and Notation}

\section{Definitions of Complexity}

\begin{definition}[supervised sample complexity]\label{supsampcomp}
Given a learning problem $(P,l,H)$ and $\epsilon, \delta >0$, we define the sample complexity
\[
m(B,H,P,\epsilon,\delta) \in \mathbb{N}
\]
of a supervised learner $B$ as the smallest natural number $k$, such that with probability at least $1-\delta$ over all possible draws of a labeled sample $S_k$, it holds that
$$R(B( S_k))- \inf\limits_{h \in H} R(h) \leq \epsilon.$$
Or in short:
\[
 m(B,H,P,\epsilon,\delta) = 
 \{\min k \in \mathbb{N} \ | \ P\left(R(B( S_k))- \inf\limits_{h \in H} R(h) \leq \epsilon) \right) \geq 1-\delta \}.
\]
Although not explicitly mentioned in the definition above, if $B$ is semi-supervised it has additional input in form of either $P(X)$, or a random draw from it. Sometimes we drop the learner $B$ from the sample complexity notation $m(B,H,P,\epsilon,\delta)$, and write either $m(H,P,\epsilon,\delta)$ or $m^{\text{\normalfont SSL}}(H,P,\epsilon,\delta)$ if there exists a supervised or semi-supervised learner respectively that achieves the sample complexity.
\end{definition}

\begin{definition}[semi-supervised sample complexity]\label{semsampcomp}
Given a learning problem $(P,l,H)$ and $\epsilon, \delta >0$ we define the sample complexity
\[
m^{\text{\normalfont SSL}}(B,H,P,\epsilon,\delta) \in \mathbb{N}
\]
of a semi-supervised learner $B$, which has information about the marginal in the form of $U \in \{U_m, P(X) \}$, is the smallest natural number $k$, such that with probability at least $1-\delta$ over all possible draws of a labeled sample $S_k$, it holds that
$$R(B( S_k,U))- \inf\limits_{h \in H} R(h) \leq \epsilon.$$
Or in short
$$
m^{\text{\normalfont SSL}}(B,H,P,\epsilon,\delta)= 
\{\min k \in \mathbb{N} \ | \ P\left(R(B( S_k),U)- \inf\limits_{h \in H} R(h) \leq \epsilon) \right) \geq 1-\delta \}.
$$
\end{definition}

We usually drop the learner $B$ from the sample complexity notation $m(B,H,P,\epsilon,\delta)$, and write either $m(H,P,\epsilon,\delta)$ or $m^{\text{\normalfont SSL}}(H,P,\epsilon,\delta)$ if there exists respectively a supervised or semi-supervised learner that achieves this sample complexity.
Similarly we drop the distribution $P$ from the notation and write $m(H,\epsilon,\delta)$ or $m^{\text{\normalfont SSL}}(H,\epsilon,\delta)$ if we can achieve this sample complexity for all distributions $P$.




\section{List of Notations Used}

\begin{table}[]
    \centering
\begin{tabular}{ll}
symbol & explanation \\ \hline
 $\mathcal{X}$  & Feature space, for example $\mathcal{X}=\mathbb{R}^n$    \\
  $\mathcal{Y}$    &  Output space. Classification: $\mathcal{Y}=\{-1,1\}$. Regression: $\mathcal{Y}=\mathbb{R}$.           \\
    $P$   &  Distribution on $\mathcal{X} \times \mathcal{Y}$            \\
    $\mathcal{P}$ & A set of distributions on $\mathcal{X} \times \mathcal{Y}$   \\
     $X,Y$  & Random variables distributed according to $P$             \\
    $P(X)$   & Marginal distribution of $P$ w.r.t to $\mathcal{X}$    \\
        $P(Y)$   & Marginal distribution of $P$ w.r.t to $\mathcal{Y}$  \\
   $D$ & Domain distribution on $\mathcal{X}$ \\
   $\mathcal{D}$ & A set of domain distributions on $\mathcal{X}$ \\
    $I_{\{\text{Boolean expression}\}}$   & Indicator function (equals 1 if expression is true and 0 else)            \\
    $ l(\hat{y},y)$  & Loss function, if not specified otherwise $l(\hat{y},y)=I_{\hat{y}=y}$         \\
    $H$   & Hypothesis class, where each $h \in H$ is a map $h:\mathcal{X} \to \mathcal{Y}$     \\
  $R(h)$     & The risk of $h \in H$. Precisely: $R(h)=\mathbb{E}_{X,Y} [ l(h(X),y)]$            \\
  $(x_i,y_i)$     & A realization of $(X,Y)$             \\
   $S_n$    &   A labeled sample set of size $n$, $S_n=((x_1,y_1),\ldots,(x_n,y_n))$          \\
 $U_m$      &  A unlabeled sample set of size $m$, usually $U_m=\{x_{n+1},\ldots,x_{n+m} \}$        \\
 $\hat{R}_n(h)=\hat{R}(h)$ & Empirical risk of h w.r.t $S_n$,  $\hat{R}(h)=\frac{1}{n}\sum_{i=1}^{n} l(h(x_i),y_i)$ \\
 $h^{\text{SSL}}$ & Model trained on $S_n$ and $U_m$ or $P(X)$, where $h^{\text{SSL}}:\mathcal{X} \to \mathcal{Y}$ \\
  $h^{\text{SL}}$ & Model trained on $S_n$, where $h^{\text{SL}}:\mathcal{X} \to \mathcal{Y}$ \\
  $m(H,\epsilon,\delta)$ & Supervised sample complexity, see Definition \ref{supsampcomp} \\
    $m^{\text{SSL}}(H,\epsilon,\delta)$ & Semi-supervised sample complexity, see Definition \ref{semsampcomp}
\end{tabular}
    \caption{List of notation.}
        \label{fulllist}
\end{table}

\vskip 0.2in

\bibliographystyle{abbrvnat}
\bibliography{SSLTHEORYBIB}

\end{document}